\documentclass{article}

\usepackage{microtype}
\usepackage{graphicx}
\usepackage{subfigure}
\usepackage{booktabs} 

\usepackage{hyperref}

\newcommand{\langpair}[2]{$\texttt{#1}\rightarrow \texttt{#2}$}
\usepackage[accepted]{icml2024}

\usepackage{amsmath}
\usepackage{amssymb}
\usepackage{mathtools}
\usepackage{amsthm}

\usepackage{amsmath,amsfonts,bm}

\def\eqref#1{equation~\ref{#1}}

\def\1{\bm{1}}

\DeclareMathAlphabet{\mathsfit}{\encodingdefault}{\sfdefault}{m}{sl}
\SetMathAlphabet{\mathsfit}{bold}{\encodingdefault}{\sfdefault}{bx}{n}

\usepackage{enumitem}
\usepackage{mathtools}
\usepackage{subfigure}
\usepackage{times}
\usepackage{latexsym}
\usepackage{xcolor, soul}
\usepackage{makecell}
\usepackage[T1]{fontenc}

\usepackage[utf8]{inputenc}

\usepackage{url}

\usepackage[english]{babel}
\usepackage{soul}

\addto\extrasenglish{

}

\usepackage[capitalize,noabbrev]{cleveref}

\theoremstyle{plain}

\theoremstyle{definition}

\theoremstyle{remark}

\usepackage[textsize=tiny]{todonotes}

\icmltitlerunning{Where does In-context Translation Happen in Large Language Models}

\begin{document}

\twocolumn[
\icmltitle{Where does In-context  Translation \\ Happen in Large Language Models?}




\begin{icmlauthorlist}
\icmlauthor{Suzanna Sia}{uni}
\icmlauthor{David Mueller}{uni}
\icmlauthor{Kevin Duh}{uni}

\end{icmlauthorlist}

\icmlaffiliation{uni}{Johns Hopkins University}

\icmlcorrespondingauthor{Suzanna Sia}{ssia1@jh.edu}

\icmlkeywords{Machine Learning, ICML}

\vskip 0.3in
]



\printAffiliationsAndNotice

\begin{abstract}
Self-supervised large language models have demonstrated the ability to perform Machine Translation (MT) via in-context learning, but little is known about where the model performs the task with respect to prompt instructions and demonstration examples.
In this work, we attempt to characterize the region where large language models transition from in-context learners to translation models.
Through a series of layer-wise context-masking experiments on \textsc{GPTNeo2.7B}, \textsc{Bloom3B}, \textsc{Llama7b} and \textsc{Llama7b-chat}, we demonstrate evidence of a "task recognition" point where the translation task is encoded into the input representations and attention to context is no longer necessary. We further observe correspondence between the low performance when masking out entire layers, and the task recognition layers. Taking advantage of this redundancy results in 45\% computational savings when prompting with 5 examples, and task recognition achieved at layer 14 / 32. Our layer-wise fine-tuning experiments indicate that the most effective layers for MT fine-tuning are the layers critical to task recognition. 
\end{abstract}

{\em In-context learning} (ICL) refers to the phenomenon in which large generative pretrained transformers (GPTs) perform tasks with no gradient updates when shown task examples or descriptions in their context~\citep{brown2020language,bommasani2021opportunities}.
While  in-context learning in GPT models appears to be generally applicable to any natural language task, to study task location, we use Machine Translation (MT) as there is little to no ambiguity in evaluating whether the model has recognised the task, since it must generate tokens in a different language. 
While in-context MT  has yet to reach parity with supervised neural MT models, it's off-the-shelf translation performance is comparatively strong and suggests a promising direction for the future of MT~\citep{hendy2023good,garcia2023unreasonable}. 
Prior work on in-context MT has focused on {\em prompt-engineering}, treating GPT models as black boxes by focusing on which examples to provide in-context \cite{moslem2023adaptive}. \citet{agrawal2022context} apply similarity-based retrieval to select in-context examples, while \citet{sia2023context} suggest a coherence-based approach.
However, these works apply surface level interventions leaving the internal mechanism of MT in GPT models largely not understood. 

In this work, we ask \textbf{where does in-context Machine Translation occur} in GPT models? We conduct an initial exploration into locating self-attention layers responsible for in-context MT in three base pre-trained and one instruction-tuned open-source GPT models .
Using causal masking over different parts of the context we demonstrate that there exists a "task-recognition" point after which attention to the context is no longer necessary (\autoref{sec:where_does_icl_happen}). The potential implications are large computational savings when the context is several times longer than the test source sentence (\autoref{sec:inference_efficiency}). 
Having identified the layers in which "task recognition" occurs, we study the extent to which subsequent layers are either \textit{redundant} or corresponding to the "task recognition" layers. Simple layer-wise masking shows that for \textsc{3B} parameter models, removing attention around the "task-recognition" layers can cause the model to fail to perform translation all-together, whereas layers towards the end of the model are much more redundant (\autoref{sec:layerwise_masking}).

Next, we observe that very lightweight fine-tuning of LoRA parameters~\citep{hu2021lora} are most effective at earlier layers of the model compared to the later ones (\autoref{sec:lora_finetuning}). This provides supports for the conjecture that earlier layers are more important for the task. 


We further investigate the extent of MT \textit{task redundancy} using differentiable $L_0$ regularisation to train discrete attention head gates (\autoref{sec:L0_train}). We find that around 10\% of the attention heads can be masked, which fundamentally differs from the literature in supervised NMT where attention heads are highly specialised for MT \citep{voita2019analyzing, michel2019sixteen, behnke2021pruning}. 

\section{Background}

\begin{figure*}[ht]
  \begin{minipage}{0.22\textwidth}
    \centering
    \includegraphics[width=\linewidth]{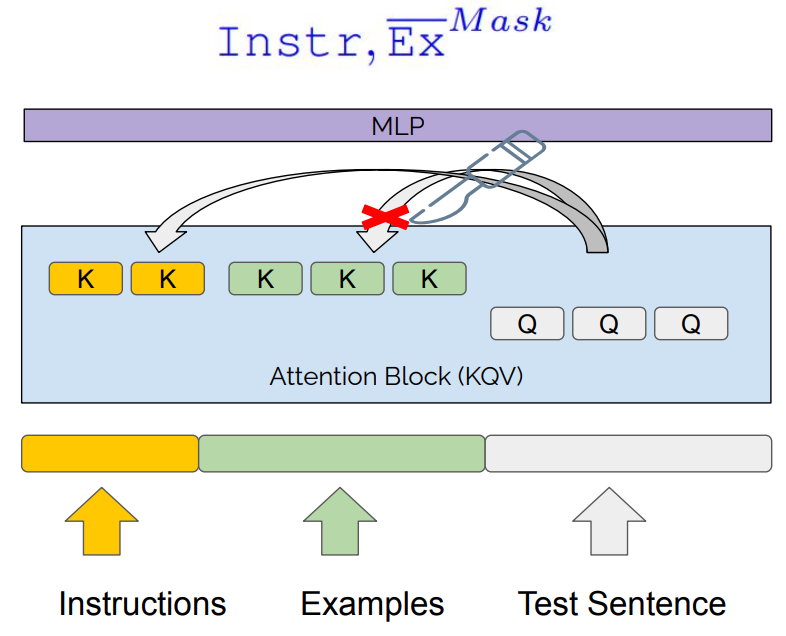}
    \label{fig:leftfigure}
  \end{minipage}\hfill
    \begin{minipage}{0.22\textwidth}
    \centering
    \includegraphics[width=\linewidth]{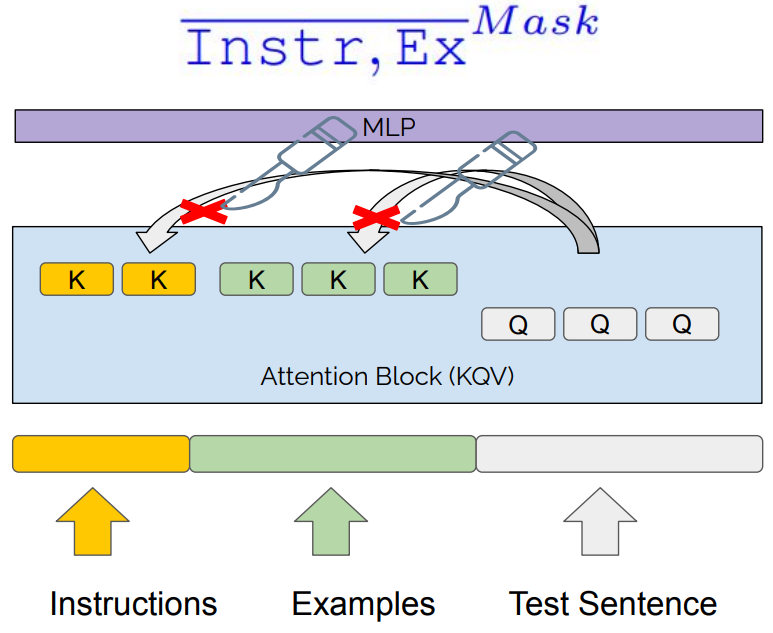}
    \label{fig:leftfigure}
  \end{minipage}\hfill
  \begin{minipage}{0.5\textwidth}
    \centering
  \resizebox{\textwidth}{!}{\begin{tabular}{l|c|c|l}

Name  & \texttt{Instr}   & \texttt{Ex}       &  \\
\toprule
\textcolor{blue}{$\overline{\texttt{Ex}}^{Mask}$} & N & \hl{Y} & \makecell[l]{\hl{Q: $\cdots$ A: $\cdots$} Q: $\cdots$ A:} \\

\midrule
\textcolor{blue}{$\texttt{Instr},\overline{\texttt{Ex}}^{Mask}$} &  Y  & \hl{Y} & Translate French to English: \hl{Q: $\cdots$ A:$\cdots$} Q: $\cdots$ A:\\

\midrule
\textcolor{blue}{$\overline{\texttt{Instr}, \texttt{Ex}}^{Mask}$}  &  \hl{Y} & \hl{Y} & \hl{Translate French to English:} \hl{Q: $\cdots$ A:$\cdots$} Q: $\cdots$ A: \\

\end{tabular}}
    \label{tab:righttable}
  \end{minipage}
  \caption{Graphical explanation of Masking the Attention over Instructions and Examples. The leftmost image has instructions and masks examples ($\texttt{Instr},\overline{\texttt{Ex}}^{Mask}$), while the right image has both instructions and examples masked ($\overline{\texttt{Instr}, \texttt{Ex}}^{Mask}$). In the interest of space we show only 2 out of 3 variants (see \autoref{fig:full_context_mask_graphics} for all variants). In the table, the overline corresponds to the yellow highlights. $N$ and $Y$ refer to absence and presence of either Instruction of Examples. $\texttt{Instr}$: Instructions and $\texttt{Ex}$: Examples.}
  \label{fig:masking_context_layers_description_wfig}
\end{figure*}

\paragraph{In-Context Learning} was first demonstrated by
\citet{brown2020language} 
 who showed that GPT-3 could be used to perform a huge variety of tasks without any task-specific parameters or training, by conditioning the model's generation on a {\it prompt} which included a few labeled examples of the task of interest.
Since then, interest in using GPT models for ICL has grown significantly, with several recent works introducing methods such as instruction-tuning~\citep{sanh2022multitask,wang-etal-2022-super} or chain-of-thought prompting~\citep{wei2022chain} to improve downstream ICL accuracy.

Ostensibly, ICL can work for nearly any task that can be defined or described in natural language, and therefore has potential for incredibly broad impact.
However, ICL can often still underperform supervised fine-tuning~\citep{bhatia2023tart}, prompting research in analyzing the mechanisms underlying ICL.
One line of work studies in-context learning with {\it linear} functions, typically linear regression, characterizing the learnability of these functions with ICL~\citep{li2023closeness, garg2022can} and even the learning algorithm a transformer uses ~\cite{akyurek2022learning, dai2023can, vonoswald2023transformers}.
A second body of work suggests that in-context learning locates {\it existing} latent concepts (tasks) which have been {\it already learnt} during pretraining~\citep{xie2021explanation,wies2023learnability}. Finally, \citet{wei2023larger} suggest that model size may change the mechanisms behind ICL from latent inference to actual learning algorithms as size increases.
Our work which focuses on Machine Translation, fits into this recent chain of work by demonstrating that there exists a point in the model's {\it layers} where the task has been located.

Many works study layers of the model as a natural unit of analysis for interpretability  \citep{hewitt2019designing, de2020decisions, pasad2021layer,durrani2022transformation, ben2022nearest, sajjad2023effect}. We highlight some of the work which is more closely related to task performance. \citet{xie2022hidden} study the layer-wise adaptability by a hidden-state variability ratio while \citet{voita2019bottom} study evolution of representations in MT-supervised transformer models. \citet{phang2021fine} studies when model layers can be skipped by feeding intermediate representations into the final output layer of a pre-trained supervised model. Our work adds to this body of work by considering the perspective of when and where layers are responsible for task location in in-context learning models. 



\paragraph{In-Context Machine Translation}
While GPT models are strong few-shot learners, their pre-training data is historically dominated by English, limiting their ability to perform translation tasks~\citep{hendy2023good}.
\citet{lin-etal-2022-shot} find that an explicitly multilingual GPT significantly outperforms traditional english models such as GPT-3, and \citet{garcia2023unreasonable} find that such models can even be competitive with supervised MT models in some settings.
However, even with explicit multilingual pre-training, in-context MT has been found to be very sensitive to the examples used~\citet{liu-etal-2022-makes} and their orders~\citet{lu-etal-2022-fantastically}.
In response, recent work focuses on how to select prompts that elicit the best downstream MT performance~\citep{agrawal2022context,sia2023context}. However, further improvement to translation with GPT models is limited by our understanding of how MT emerges in GPT models.
Our work directly analyses when, in layer representations, a GPT model becomes a translation model via in-context learning, and how that may inform decisions around parameter tuning and redundancy.

\section{Data and Settings}

\paragraph{Models}
We use \textsc{GPTNeo2.7B} \citep{gpt-neo}, \textsc{Bloom}3B \citep{scao2022bloom}, \textsc{Llama}7B and \textsc{Llama}7B-chat \citep{touvron2023llama}, the instruction-tuned variant, in all of our experiments. \textsc{GPTNeo2.7B} has 32 layers and 20 heads, \textsc{Bloom3B} has 30 layers and 32 heads, while \textsc{llama}7B has 32 layers and 32 heads. The checkpoints we use are from the transformers library~\citep{wolf2019huggingface}.

\textsc{GPTNeo} was trained on The PILE \citep{gao2020pile}, an 825GB text dataset which consists of roughly 98\% English data. Despite being mostly monolingual, The PILE contains Europarl which \textsc{GPTNeo} was trained on at a document level (rather than a sentence level).
Conversely, \textsc{Bloom} was trained on the ROOTS corpus \citep{laurenccon2022bigscience}, a composite collection of 498 datasets that were explicitly selected to be multilingual, representing 46 natural languages and 13 programming languages.  \textsc{Llama} training data consists primarily of common crawl, C4, wikipedia, stackexchange as major sources. To our knowledge, there has not been any reports of sentence level parallel corpora in the training datasets of these models.

\paragraph{Data}
We test our models using FLORES \citep{flores} $\texttt{en}\!\leftrightarrow\! \texttt{fr}$ which we report in the main paper, and a small study on extending \autoref{sec:where_does_icl_happen} to $\texttt{en}\!\leftrightarrow\! \texttt{pt}$ in the Appendix. Prompt examples are drawn from the development set. 
We evaluate the generations using BLEU scores, following the implementation from \citet{post2018call}. 

\paragraph{Prompt Format}

Our prompts may consist of instructions, examples, both, or none. Importantly, we adopt \textit{neutral} delimiters, "Q:" and "A:" to separate the prompt and the start of machine generated text. This ensures that the models do not have any information from the delimiters on what the task is. \footnote{In an earlier exploration, we found that supplying the model with language indicators only, e.g., "English:", "French:" was sufficient for strong models (llama7b, llama7b-chat) to perform the task without seeing any instructions or examples in the context.}

When no natural language instructions are used the model input will be \texttt{Q: \{source\_sentence\} A:} 
Instructions are given in natural language and take the form: \texttt{Translate from \{L1\} to \{L2\}: Q: \{source\_sentence\} A:}, where \texttt{L1 = English} and \texttt{L2 = French} if the source and target languages are English and French respectively. Examples are given after instructions, and similarly delimited by Q: and A:. See Appendix: \autoref{tab:format_example} for an example.


\section{Where does In-context MT happen?} 
\label{sec:where_does_icl_happen}

\subsection{Layer-from Context Masking}

In-context learning differs from task-specific supervised learning in that, during test time, the desired task must be identified from the context first, then executed. At what stage in the feed-forward computation does a GPT-style model transition from an in-context learner to a translation model? To explore this question, we introduce {\it layer-from context-masking} which masks out all attention weights to the context (instructions or prompts) \textit{from} a certain layer onwards 
 (see \autoref{fig:masking_context_layers_description_wfig} for a graphical description). 

For Causal Decoder-only Transformer Language Models, given each position $i$, the Attention weight $\alpha_{ij}$ over context positions $j, j<i$ can be computed by a $\alpha_{ij}=\mathrm{softmax}(\frac{QK^T}{\sqrt{d_k}})_{ij}$. Each element in $(QK^T)$ is the dot product between a query vector and key vector $q_i \cdot k_j$, where $q_i=W_q x_i, k_j=W_kx_j$ for trained weight matrices $W_k$ and $W_q$.\footnote{Readers should note that there is a $W_k$ and $W_q$ weight matrix for each layer and each attention head, but we omit the notation on this for readability.} We apply the attention mask over the context so that the attention score is $(q_i \cdot k_j) + m(j, \mathbf{u})$. Here $\mathbf{u}$ is the context that we wish to mask, and $m(j, \mathbf{u}) = 
\begin{cases}
    0 & \text{if } x_j \notin \mathbf{u} \\
    -\infty & \text{if } x_j \in \mathbf{u}
\end{cases}
$

All masks operate from the $j$-th layer ($\ell_j)$ \textit{onwards}, i.e. masking from $\ell_{20}$ means causally masking out attention to all context positions from $\ell_{20:n_{\ell}}$, where $n_{\ell}$ is the total number of layers. To construct Fig 2, we increment $\ell$ from $1$ to $n_{\ell}$ and apply the set of masks $\{m(j,\textbf{u})\}^{\ell:n_\ell}$ in each experiment and observe the performance of the model. 

Under this causal masking treatment masking from layer $\ell$, the model must rely on the representations of the target input sentence from layer $\ell+1$ {\it only} to complete the task; if the target sentence representations do not already encode the target task (translation into a specific language) then the model will fail to generate translations. 

In other words, the goal is to characterise where the model has "located” the task of translation. In all experiments we mask the examples provided in the context, but to control for the effect of semantic instructions, we ablate over different treatments of the instructions by removing instructions entirely ($\overline{\texttt{Ex}}^{Mask}$), leaving them unmasked ($\texttt{Instr}\overline{\texttt{Ex}}^{Mask}$),  or masking them together with the examples ($\overline{\texttt{Instr}\texttt{Ex}}^{Mask}$). The overline notation indicates the context which are masking over. Also see \autoref{fig:masking_context_layers_description_wfig}.


\subsection{Results}

We discuss the central findings of the paper: \textbf{Models do not need to maintain attention over all of the context across every layer to perform the task.}

\begin{figure*}[!t]
\centering
    \includegraphics[width=\columnwidth,trim=0 0 0 0]{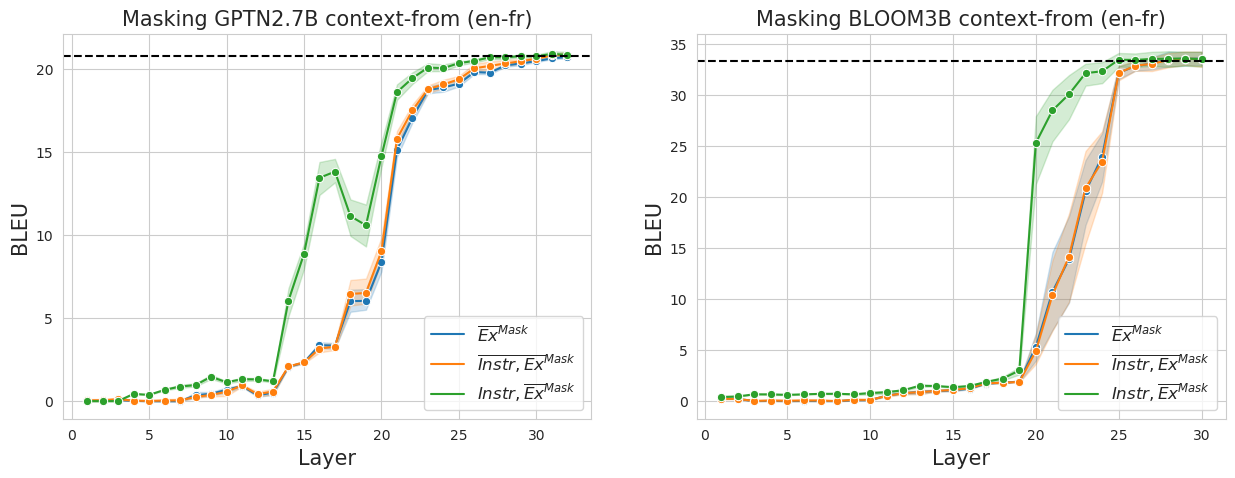}
    \hfill
    \includegraphics[width=\columnwidth,trim=0 0 0 0]{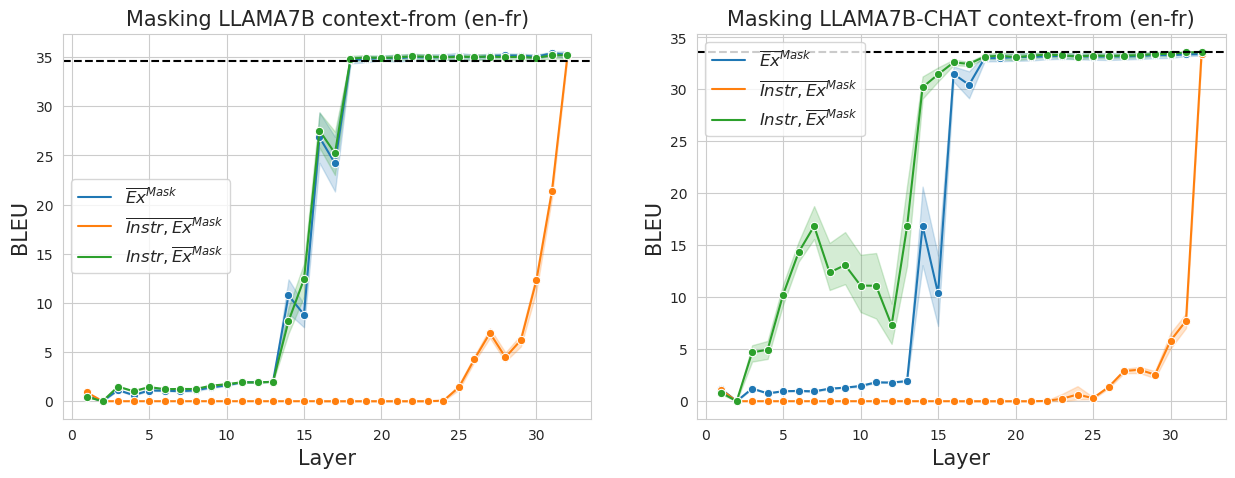}
    \hfill
    \includegraphics[width=\columnwidth,trim=0 0 0 0]{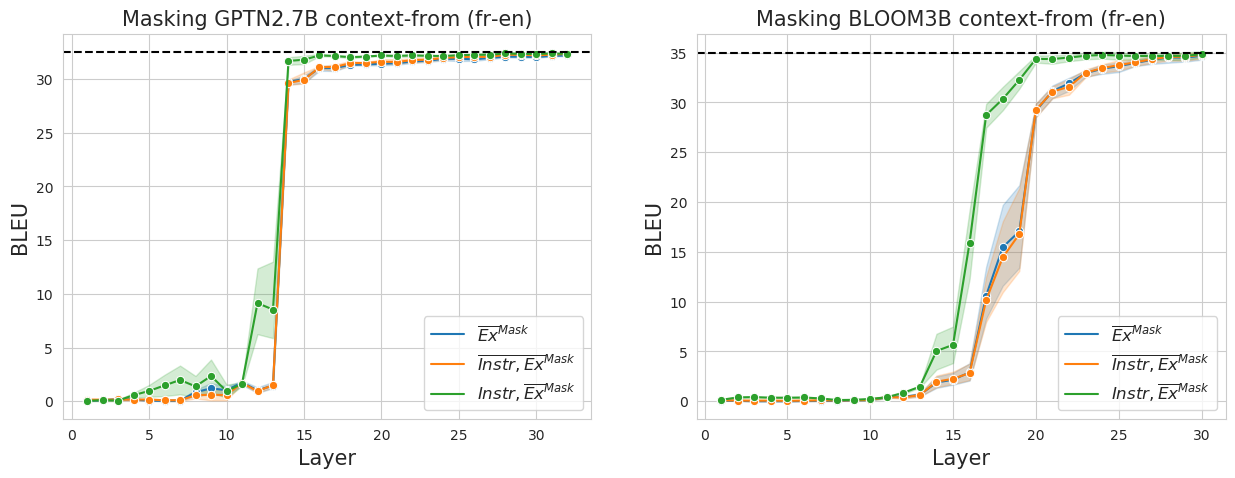}
    \includegraphics[width=\columnwidth,trim=0 0 0 0]{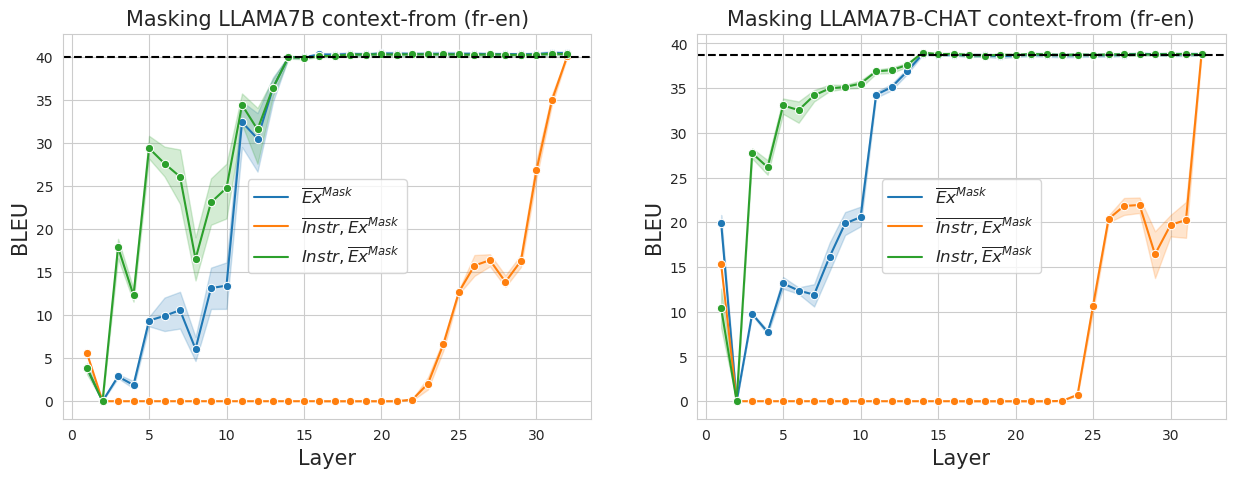}
    \caption{\textit{Layer-from context-masking experiments} for GPTNeo2.7B, BLOOM3B, Llama7b, Llama7b-chat on $\texttt{en}\!\leftrightarrow\! \texttt{fr}$. The graphs show translation performance when masking contexts from the  $j^{\textrm{th}}$ layer onwards. Different lines indicate different treatments of the instruction, as described in \autoref{fig:masking_context_layers_description_wfig}. The dashed black line is the performance when shown both examples and instructions without masking.}
    \label{fig:context_mask_fig1}
    
\end{figure*}

In all models we observe that when applying masking from $\{m(j, \textbf{u})\}^{\ell:n_\ell}$ over the context, performance plateaus before the final layer, i.e., when $\ell=n_\ell$. The results of our experiment for $\texttt{en}\!\rightarrow\! \texttt{fr}$ and $\texttt{fr}\!\rightarrow\! \texttt{en}$ are shown in \autoref{fig:context_mask_fig1}, and additional experiments for GPTNeo and Bloom on $\texttt{en}\!\rightarrow\! \texttt{pt}$ and $\texttt{pt}\!\rightarrow\! \texttt{en}$ are shown in \autoref{sec:appendix-spanish}. 

Different models reach this plateau point at different layers. In \textsc{GPTNeo} this point occurs around layer 25, in \textsc{BLOOM} this point occurs around layer 15-20, and in \textsc{Llama} models this occurs around layer 13-15. As English is the dominant language, as expected models can successfully perform translation into English upon earlier layers of masking, than translation out of English.

At this point, the models benefits only marginally, if at all, from attending to the context, suggesting most of the task "location" has already occurred.

\textbf{There exists critical layers for task location}. Prior to the task recognition point, around the middle layers of the models, moving the context mask up a layer results in a significant increase to performance. We consider these critical layers, as instead of a gradual increase in performance, we observe we can observe very steep jumps of over $20$ bleu points across the different models. We conjecture that the model is locating the correct task during processing in these middle layers, after which the context is no longer necessary to perform the translation task.

 Overall, our findings suggest a 3-phase process to in-context learning: in the first phase, moving the mask up makes little difference in performance, which is close to 0. This suggests that the context has not influenced task location at all. In the second phase, shifting the mask upwards makes a large difference in MT performance, suggesting that the model has started to locate the task but can improve significantly with more processing of the context. Finally, in the third phase, shifting the mask upwards again has little-to-no effect on MT performance, suggesting that the model has fully recognized the task as translation and no longer requires the context to interpret the task. 

 We provide further observations and ablations in the following sections.

\subsection{Instruction-tuned vs Non-instruction Tuned Models}

When comparing non-instruction tuned vs instruction-tuned \textsc{Llama7b} models, we do not observe any noticeable difference in where performance plateaus, i.e., where the model no longer requires attention over the context. This occurs around layers $18$ for both \textsc{Llama} models in $\texttt{en}\rightarrow \texttt{fr}$ and around layer $14$ for $\texttt{fr} \rightarrow \texttt{en}$. The main difference is that instruction-tuned model is able to achieve better performance in the earlier layers for the setting where instructions are present and examples are masked ($\texttt{Instr},\overline{ \texttt{Ex}}^{Mask}$). This is to be expected as these models are tuned towards following instructions. 

Overall we find that the observation of task recognition layers and a task recognition point is present across both non-instruction tuned and instruction tuned models, and that this presents itself similarly in both types of models.

\subsection{The Role of Instructions vs Examples}

In separate experiments, we found that when shown only instructions and no examples, \textsc{GPTNeo} and \textsc{Bloom} models are unable to translate, and their performance is nearly at 0 BLEU Score. For \textsc{GPTNeo} and \textsc{Bloom} we see that the behavior of the model is similar when no instructions are present  ($\overline{\texttt{Ex}}^{Mask}$) and when instructions are masked ($\overline{\texttt{Instr,Ex}}^{Mask}$). However, if the model is given complete access to instructions ($\texttt{Instr}\overline{\texttt{Ex}}^{Mask}$), it can use the intermediate processing of examples to reach baseline performance earlier. 

\subsection{Attention to the Context vs Attention to the Input}

\begin{figure*}[!h]
\centering
\includegraphics[width=\columnwidth]{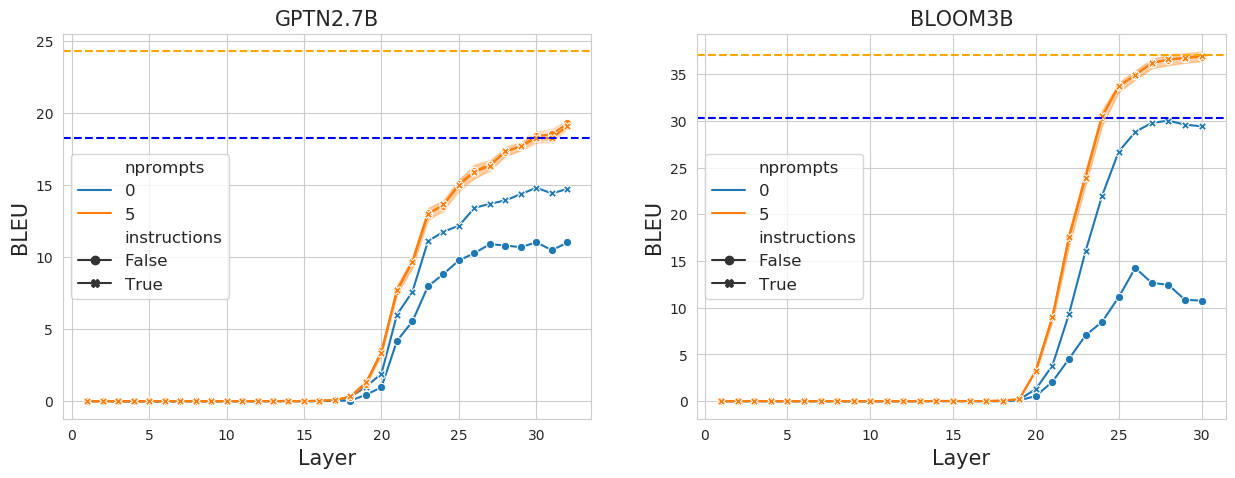}
\includegraphics[width=\columnwidth]
{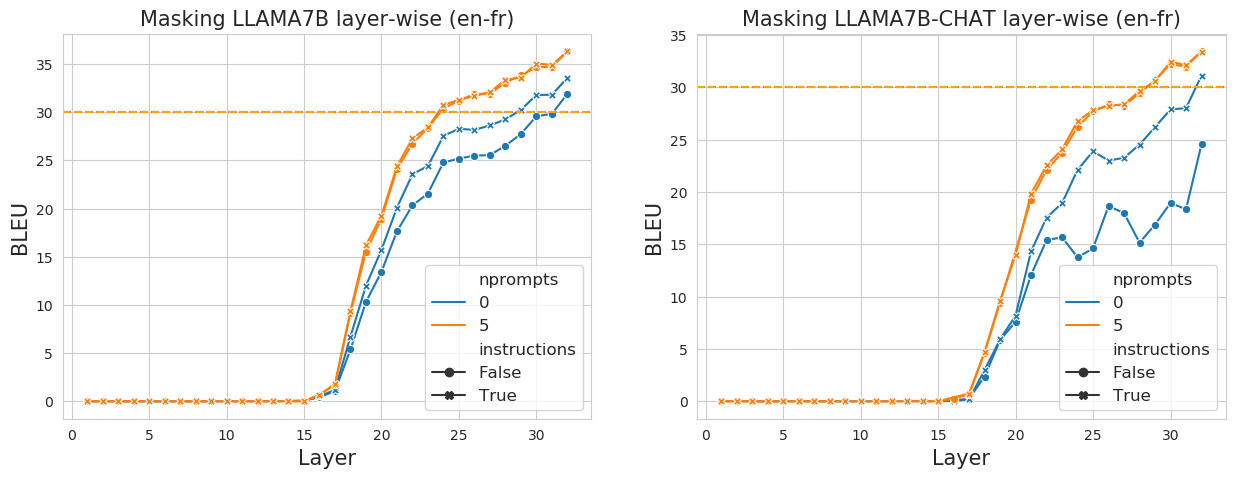}
\caption{\textit{Layer-from} experiments for \textsc{GPTNeo}2.7B, \textsc{Bloom}3B, \textsc{Llama} and \textsc{Llama7b-chat} on $\texttt{en}\rightarrow \texttt{fr}$ when masking out from layer $j$ onwards. Orange and blue dashed lines refer to the baselines (without masking) of 0 and 5 prompts with instructions. In view of the smaller models failure to translate at all under the format Q: A: with no examples, we adopt "English:", "French:" as delimiters instead of QA in generating this figure.}
\label{fig:mask_from_layer}
\end{figure*}

One possible explanation for the results in \autoref{fig:context_mask_fig1} is that, rather than identifying the point at which the task is recognized, we have identified the point at which the model no longer requires attending to {\it any} other input tokens.
To explore this, we run experiments in the \langpair{en}{fr} direction where we mask attention to {\it all inputs} from a certain layer onwards. This does not include masking over the text the model has generated.

We plot the results in \autoref{fig:mask_from_layer};
we find that for all models, the layer at which attention can be fully removed is much higher than the layer at which we can remove attention to the context.
For \textsc{GPTNeo} and \textsc{Llama}, translation performance is never comparable to the baseline with no masking.
Conversely, when masking only the context, translation performance improves as early as layer $10$ and plateaus at the no-mask baseline much earlier.
This supports the interpretation that the curves we observe in \autoref{fig:context_mask_fig1} are due to the model still requiring attention to the source sentence input.

\begin{figure*}[!t]
\centering
    \includegraphics[width=\columnwidth,trim=0 0 0 0]{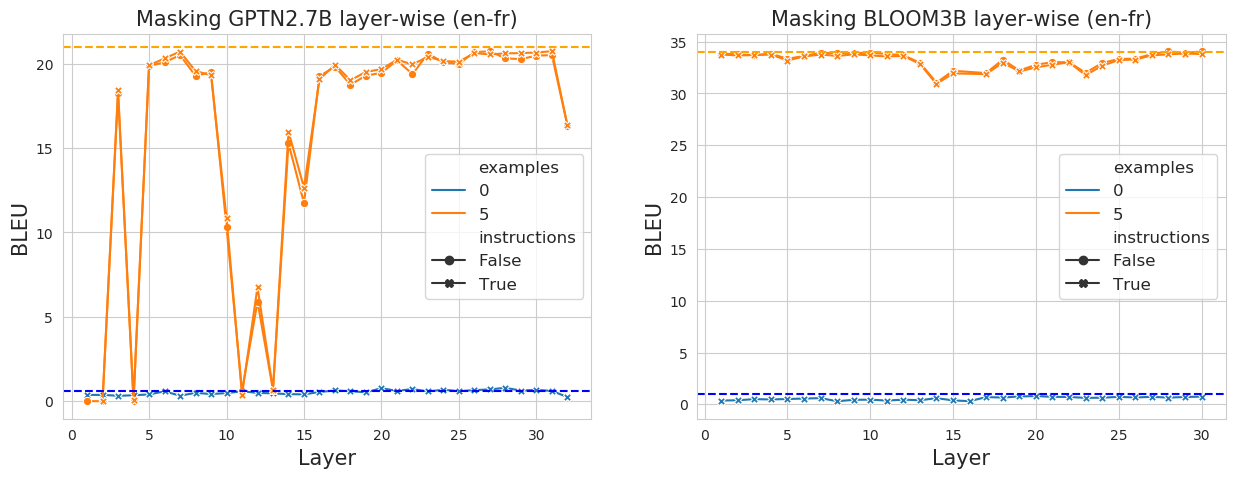}
    \hfill
    \includegraphics[width=\columnwidth,trim=0 0 0 0]{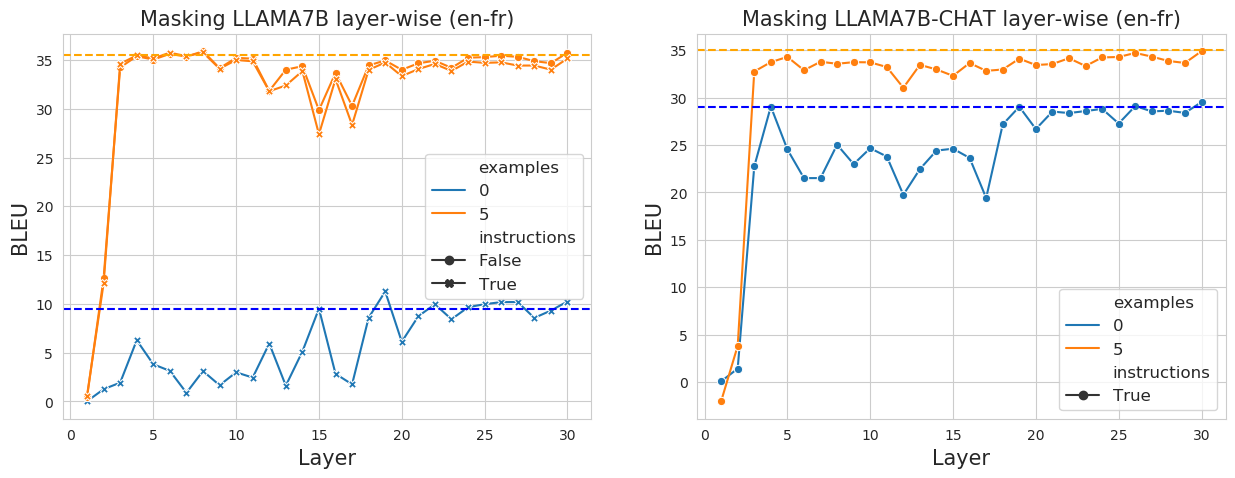}
    \hfill
    \includegraphics[width=\columnwidth,trim=0 0 0 0]{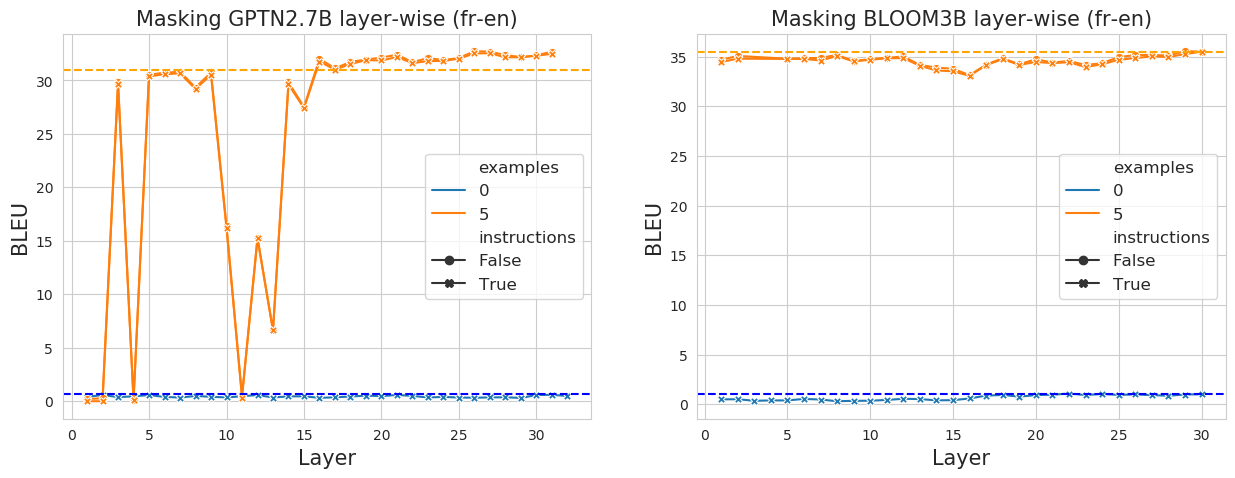}
    \includegraphics[width=\columnwidth,trim=0 0 0 0]{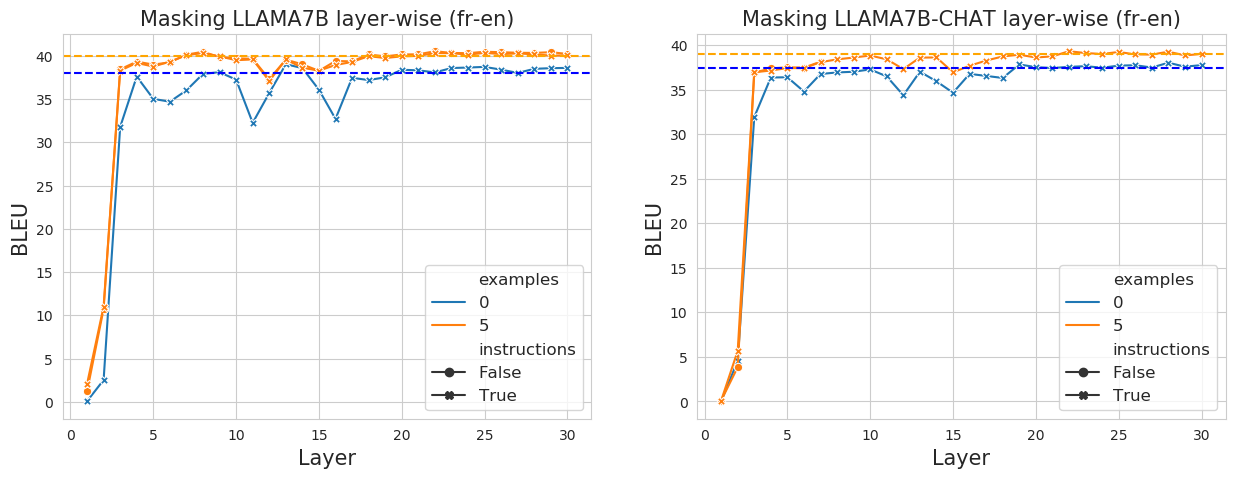}

    \caption{\textit{Layer-wise masking} of self-attention heads for \textsc{GPTNeo}2.7B, \textsc{Bloom}3B, \textsc{Llama} and \textsc{Llama-chat} on $\texttt{en}\leftrightarrow {fr}$. The orange and blue dotted lines refer to the baselines (without masking) of 0 and 5 prompts with instructions. We observe critical layers near the middle and redundant layers towards the end of the model.}

    \label{fig:mask_out_layer_gptn}
\end{figure*}

\section{Characterising Redundancy in Layers}
\label{sec:redundancy}

Recently, \citet{sajjad2023effect} found that many layers in pre-trained transformers can be dropped with little harm to downstream tasks; moreover, it is well known neural MT transformer models are known have several redundant heads which are not necessary during test time~\citep{voita2019analyzing,michel2019sixteen,behnke2021pruning}. However, it is not clear if the same trends hold for {\it in-context MT} models, and how that redundancy is related to task location versus task execution.

We study the contributions of individual attention-layers by performing a simple {\it layer-wise} masking of all self-attention heads for a single layer.
When we mask layer $j$, we are masking the {\it attention mechanism} of layer $j$, that is the MLP of layer $j$ acts directly on the output of layer $j-1$, rather than the output of the attention-head of layer $j$.
Doing so allows us to study how {\it critical} each layer is, where \textit{critical layers} is loosely defined as those that have a large negative impact when masked. 

We plot results for each layer all models, using the three combinations of \{0 examples, no instructions\}, \{5 examples, instructions\}, \{5 examples, no instructions\} in \autoref{fig:mask_out_layer_gptn}.\footnote{The combination of \{0 examples, no instructions\} is not meaningful as the model only receives "Q: <source sentence> A:" as the input and is not expected to do the translation task.}

\subsection{Are ``Critical" Layers Task Locating Layers?}

\label{sec:layerwise_masking}

In \autoref{sec:where_does_icl_happen}, we observed that there are layers for task location. In this section, we observe evidence that there are critical layers which correspond to the task locating layers, providing support for our earlier observations. 

 For instance for \textsc{Llama7b} \langpair{en}{fr}, even in the scenarios when examples are provided, we can see a dip in performance around layer $15$ to $18$. Refering back to \autoref{fig:context_mask_fig1}, we see that this is where most of the task location with large jumps in performance had occurred. 

For GPTNeo, we obseve a large set of contiguous layers which significantly decrease performance at around layer $10$ to $15$. This also corresponds to where most of the task location (large jumps in performance) had occurred for this model in \autoref{fig:context_mask_fig1}. 

We note that the critical layers in different models have varying degrees of severity. It is not immediately clear why \textsc{GPTNeo} has such critical layers and suffers compared to the other models, although we note that this is unlikely to be due to size or model architecture  as \textsc{Bloom} is also around the same size as \textsc{GPTNeo} and performs more similarly to \textsc{Llama}. We suspect that it could be due to training data or some other factor related to the training dynamics but leave this for future work.

With regard to redundancy, we find that layers can be more safely removed towards the end without a noticeable loss in performance. We observe that for the less stable models, the model achieves close to baseline performance by layer-wise masking from $\ell_{15}$ for \textsc{GPTNeo}, $\ell_{26}$ for \textsc{Bloom} and $\ell_{20}$ for \textsc{Llama}. This suggests that these later layers contain redundancy for translation.

Overall, observing redundancy in layers is not suprising, and our main contribution is characterising the differences between redundant and critical layers. To explain why models can have redundant layers, we refer to \citet{clark2019does} who identify a phenomena where attention heads attend almost exclusively to delimiter and separator
tokens such as \texttt{[SEP]}, periods and commas. This is thought to act as a ``no-op" as the value of such tokens in changing the current hidden representation is very small. Note that it is then possible to mask entire Transformer layers and still achieve a sensible output due to residual connections in the Transformer architecture at every layer.

\section{Inference Efficiency}
\label{sec:inference_efficiency}
Speeding up transformer inference is of great interest to the community \cite{fournier2023practical}. We highlight the potential of speeding up inference time as a direct consequence of identifying where task recognition occurs in the model and redundancy of self-attention processing. Our results indicate that we can achieve significant speedups in inference by removing the processing of context-tokens all-together after a certain point in the model, with little to no impact on downstream performance. 

Let $\ell_r$ be the $r^{\textrm{th}}$ layer where we can mask out the attention of the context across subsequent layers and match the ``ceiling" performance. Let $k$ be the number of prompt examples, where each example consists of a pair of parallel sentences. Then, for a model with $n_\ell$ layers, the amount of processing in terms of speed and memory saved is approximately $(n_\ell - r) / n_\ell \times  (k / k + 1)$. 

Using the example of \textsc{Llama7b} ($32$ layers), we see from \autoref{fig:context_mask_fig1} that the model is very close to it's ceiling score after processing the examples at layer 14 ($\ell=14$). If we no longer need to process examples after $\ell=14$, \textbf{under a prompt size of $5$ the savings are approximately 45\%.} 

For instruction-tuned models which are typically deployed in production, even if we assume that no examples are provided, savings can be non-trivial as very long-form instructions are typically provided to the model in an attempt to control it's behavior (prompt engineering).

\section{Further Analysis}
In the following sections, we focus on \textsc{GPTNeo} and \textsc{Bloom} to conduct deeper analysis on the main phenomena presented in the paper.

\subsection{Does the Number of Prompts Affect Task Recognition?}

\label{sec:layerfrom_context_masking_nprompts}

\begin{figure}[!t]
\centering
    \includegraphics[width=0.9\columnwidth,trim=0 0 0 50]{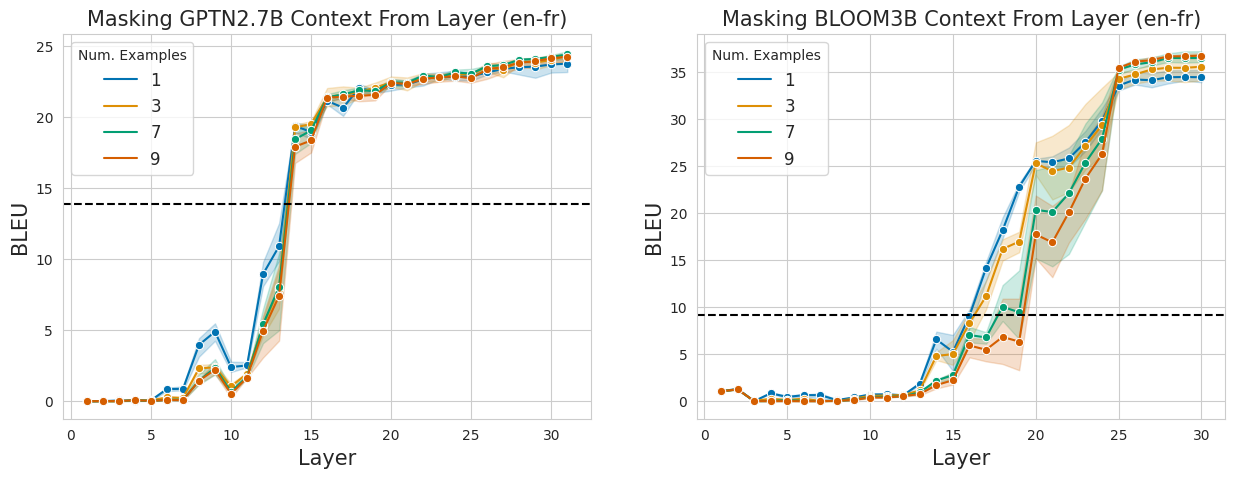}    
    \caption{\textit{Layer-from context-masking experiments} for GPTNeo and BLOOM on $\texttt{en}\!\rightarrow\! \texttt{fr}$ investigating number of examples in the $\overline{\texttt{Ex}}^{Mask}$ mask setting (described in \autoref{fig:context_mask_graphics}). The dashed black line refers to no instructions and no examples. }
    \label{fig:mask_context_nprompts}
\end{figure}

In \autoref{sec:where_does_icl_happen} we study context-masking with a fixed number of prompts.
However, it is not clear if the number of prompts affects how fast, layer-wise, the model is able to recognize the task.
We plot these results for $\texttt{en}\!\rightarrow\! \texttt{fr}$ in \autoref{fig:mask_context_nprompts}, for both \textsc{GPTNeo} and \textsc{BLOOM}.
In general, we find that the number of prompt examples has little effect on which layer the task is recognized at.
While there is some variation in performance when the context is masked around the middle layers of the model, the final performance plateau occurs at the same layer regardless of the number of prompts.

\subsection{The Adaptability of Task Layers}
\label{sec:lora_finetuning}

\begin{figure}[!t]
    \centering
\includegraphics[width=\columnwidth]{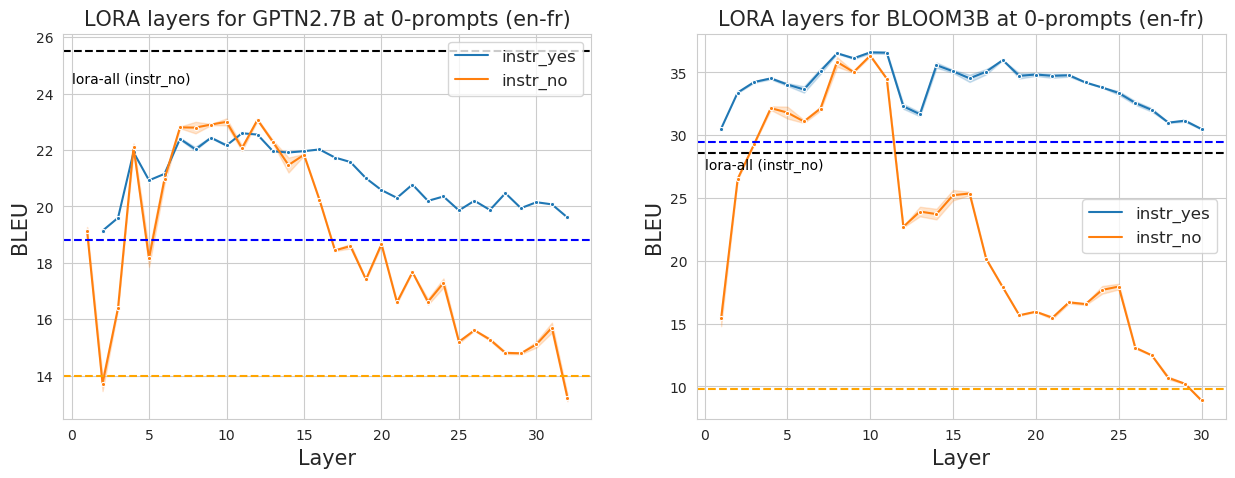}
\hfill
\includegraphics[width=\columnwidth]{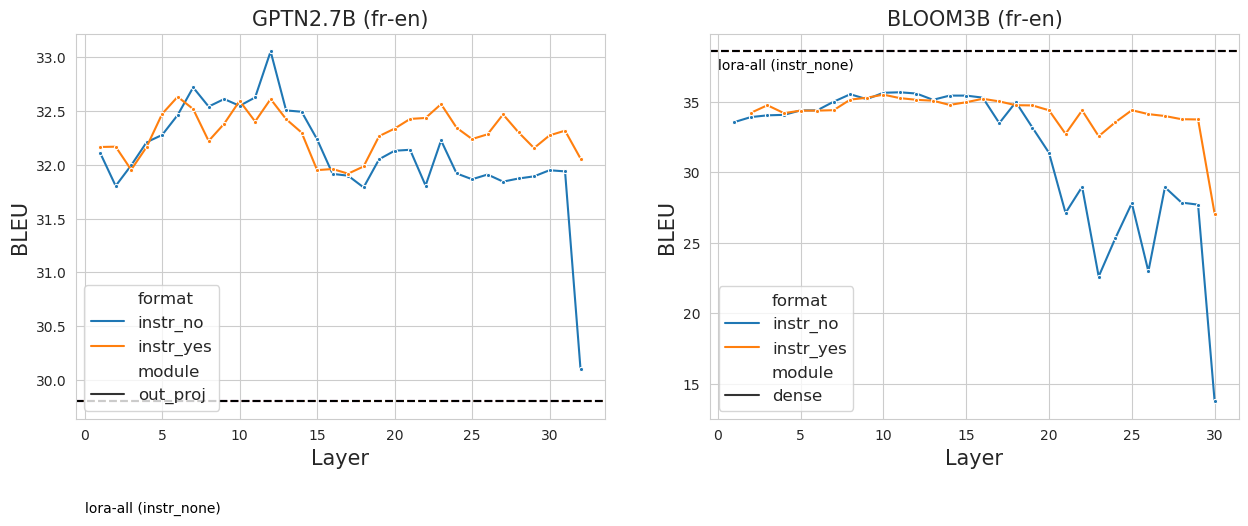}
\caption{Performance of no-instructions trained Lora layers for GPTNeo and BLOOM on \texttt{en}$\leftrightarrow$\texttt{fr}. The dashed black line refers to training of all layers together, while the orange  (test without instructions) and blue (test with instructions) dashed lines refers to no training. The layers which are most amenable to lightweight fine-tuning occur in the earlier layers before the "task recognition" point.}
\label{fig:lora_eval}
\end{figure}

Intuitively, the layers prior to "task recognition" should contain information about locating the MT task. To test this intuition, we further explore the adaptability of these layers by lightweight fine-tuning experiments. We trained a single Low-rank Adaptation matrix (LoRA; \citet{hu2021lora}) for each layer of the output projection while keeping the rest of the network frozen.\footnote{We also experimented with the training separate Key, Query and Value LoRA Layers but found this to be less effective.} 
The model was shown parallel sentences as input, and layers were trained with no explicit translation instructions. We split the dev set of FLORES into $800$ training examples and $200$ dev examples. Note that this setup is designed to tune the layers for task location. It is highly unlikely that the model can learn translation knowledge from this small amount of supervision. The LoRA layers were trained for up to $50$ epochs with early stopping patience$=5$ and threshold$=0.001$, with $\alpha=32, r=32$ and dropout$=0.1$. The cross-entropy loss was computed only on the target sentence (see \autoref{sec:masked_training} for details) and we used the best checkpoint on the 200 held out dev examples for evaluation.

We show the results of this experiment in \autoref{fig:lora_eval}; while each layer can be trained to perform better than no fine-tuning at all, tuning different layers have different impacts on performance. In particular, we find that high performing layers occur at the earlier to middle parts of the network, with the peak often occurring near the start of the "task-locating" layers from \autoref{sec:where_does_icl_happen}. In contrast to common fine-tuning wisdom, additional tuning on the later layers has a much smaller impact on final performance for $\texttt{en}\rightarrow\texttt{fr}$. 

\subsection{Are There Specialised Attention Heads?}
\label{sec:L0_main}
In \autoref{sec:where_does_icl_happen}, we found that the earlier part of the model is critical for \textit{task location} from the prompt context, and in \autoref{sec:layerwise_masking} we found both critical and redundant layers to the MT task. In this section, we increase the level of granularity to that of attention heads instead of layers.

A well established finding for supervised encoder-decoder MT models, is that up to 90\% of the attention heads can be pruned while minimising fall in translation performance \citep{voita2019analyzing, behnke2020losing, michel2019sixteen}. We note that asking about the extent of pruning is a slightly ill-formed research question, as it depends on the type of pruning technique used. However broad trends of highly prunable models have been observed in the supervised MT paradigm. In the in-context paradigm, there is no explicit supervision. Thus it is not clear if the task knowledge is spread across a much larger number of attention heads, or similarly specialised to a few heads. For instance, \citet{bansal-etal-2023-rethinking} studied attention-head importance for a broader set of ICL tasks, finding that the most important heads for ICL occur in the middle layers of the model.

\subsection{Training Attention Head Gates with $L_0$ regularisation}
For a scalable approach to pruning, we opt to train self-attention head gates following \citet{voita2019analyzing} using the technique of differentiable $L_0$ regularization \citep{louizos2017learning}.
Let the attention head gates $g \in \mathbb{R}^{n_h \times n_{\ell}}$ be a set of trainable parameters, where $n_h$ is the number of attention heads per layer, and $n_{\ell}$ is the number of layers. Let the original output of each attention head be $v_j$, gated outputs $\tilde{v}_j$ are obtained by elementwise multiplication of the gate value $g_j$, i.e., $\tilde{v}_j = g_j \odot v_j$. For $\{(x, y)\}^n$ source sentence $(x)$ and target sentence $(y)$ training pairs, a model $f$ and loss function $\mathcal{L}$, $L_p$ regularisation adds a $\lambda$ weighted penalty associated with the complexity of the parameters. \footnote{$L_2$ regularisation has the effect of reducing the magnitude of all $g$, $L_1$ regularisation has the effect of reducing the magnitude of several attention heads to a very small value (but not exactly 0), while $L_0$ regularisation has the effect of driving $g$ values to exactly 0.}
The $L_0$ loss is non-differentiable as it involves raw counts of parameters. As a work around, $g$ can be approximated with random variables drawn from a Binary concrete distribution \citep{maddison2016concrete, jang2016categorical}.\footnote{The class of Concrete distributions was invented to work around the problem of automatic differentiation of stochastic computation graphs.} We refer the reader to \citet{louizos2017learning} for the relevant technical exposition. Details of training  are provided in \autoref{sec:L0_train_details}.

\begin{figure}[!t]
    \centering
\includegraphics[width=0.9\columnwidth]{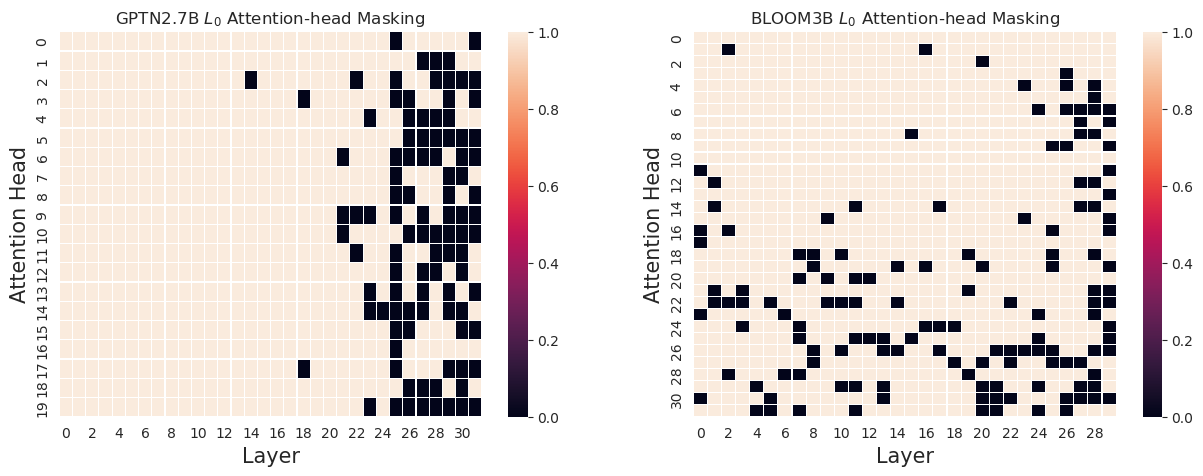}
    \caption{Visualisation of attention head masks for GPTNeo and BLOOM, learned with $L_0 (\lambda=0.01)$ regularisation under a \texttt{0-prompt train} scheme in $\texttt{en}\rightarrow\texttt{fr}$. A value of $0$ (in black) indicates that the attention head is effectively masked out by the trained attention gate. Around 10\% of attention heads are masked out i.e., redundant, with a majority of them occuring at the later layers for GPTNeo and distributed across layers for BLOOM. $\texttt{fr}\rightarrow\texttt{en}$ is availble in \autoref{sec:appendix_L0}}
    \label{fig:grid_enfr}
\end{figure}

\subsection{Studying Redundancy via Compression}
\label{sec:L0_train}

We noted that \textsc{GPTNeo} has some critical differences from \textsc{BLOOM} and \textsc{Llama} in terms of having critical layers (see \autoref{sec:layerwise_masking}). To what extent are there specialised attention heads for MT in the GPT-style models? If there were specialised heads, we would expect the model to be highly compressable/prunable to a select few heads. We plot a grid map of learned attention gate values for \texttt{en} $\rightarrow$ \texttt{fr}, where 0 indicates that the head is masked out (\autoref{fig:grid_enfr}). We find that most of the masked heads are distributed at the later layers for GPTNeo and are distributed across layers for BLOOM. This appears consistent with \autoref{sec:layerwise_masking}’s observations that redundancy is more focused at certain layers in GPTNeo, and more spread out across the layers for Bloom.

In addition, we note that there are no "few" specialised heads, which directly contrasts with the literature on compression in supervised MT models \citep{voita2019analyzing, michel2019sixteen}. Potential reasons for this difference might include data distribution and model architecture, or cross-entropy loss associated with task tuning for MT vs non-specific training on large corpora. We leave this as an open question for future work.

\section{Conclusion}
We demonstrate evidence that In-context Causal Decoder models locate the translation task at a specific layers during forward inference. To study this, we introduced causal masking of self-attention over the context from layer $\ell$ onwards (\autoref{sec:where_does_icl_happen}). The findings generalise across models of different sizes and in both non instruction-tuned and instruction-tuned models. We further identify certain layers as task critical, and show that this corresponds to the task recognition point of the model  (\autoref{sec:layerwise_masking}) and is not influenced by increasing number of examples (\autoref{sec:layerfrom_context_masking_nprompts}) shown to the models.

Our central finding that models do not need to maintain attention over all of the context across every layer has direct implications for inference efficiency of transformers, with estimated up to 45\% cost-savings for llama model with 5 examples (\autoref{sec:inference_efficiency}).

Contrary to common fine-tuning wisdom, we show that it is sometimes beneficial to target middle layers for fine-tuning the model which could be associated with task recognition ( \autoref{sec:lora_finetuning}). Finally, we trained attention head gates using differentiable $L_0$ regularisation (\autoref{sec:L0_main}), and found that around 10\% of attention heads can be masked. These are mostly distributed across the later layers of the model, providing some support for the idea that later layers are redundant but layers are responsible for locating the translation task. Although we have characterised this phenomena using the example of translation we believe that the broad findings are likely to generalise to different tasks.
\paragraph{Limitations and Future Work}
We have conducted extensive investigations focusing on the task of translation on a high-resource language pair, with a small extension to $\texttt{en}\leftrightarrow \texttt{pt}$.  In future work, we hope to extend this analysis to other sequence or classification tasks as well as \textit{true} novel tasks. 

\paragraph{Reproducibility}
The MT dataset that we use, FLORES \citep{flores}, is fully open-source and well-known in the community. Models are open-source and freely available on Huggingface \citep{wolf2019huggingface}. We used models of "reasonable" size (3B and 7B parameters) that can be run with consumer grade GPUs, making our reproducible to most academic institutions. Code to reproduce all the experiments will be made available subsequently.  

\paragraph{Impact Statement (Ethics and Societal Consequences)}
There are no known ethical concerns as these are exploratory studies on open-source LLMs. 


\subsubsection*{Acknowledgments}
We would like to thank Daniel Kashabi and Marc Marone for feedback on earlier drafts.

\clearpage
\bibliography{icml}
\bibliographystyle{icml2024}
\clearpage
\appendix
\section{Appendix}
\label{sec:appendix}

\subsection{Graphical View of Context Masking Experiments}
\label{fig:full_context_mask_graphics}
\begin{figure*}[!t]
\centering
    \includegraphics[width=0.6\columnwidth,trim=0 0 0 0]{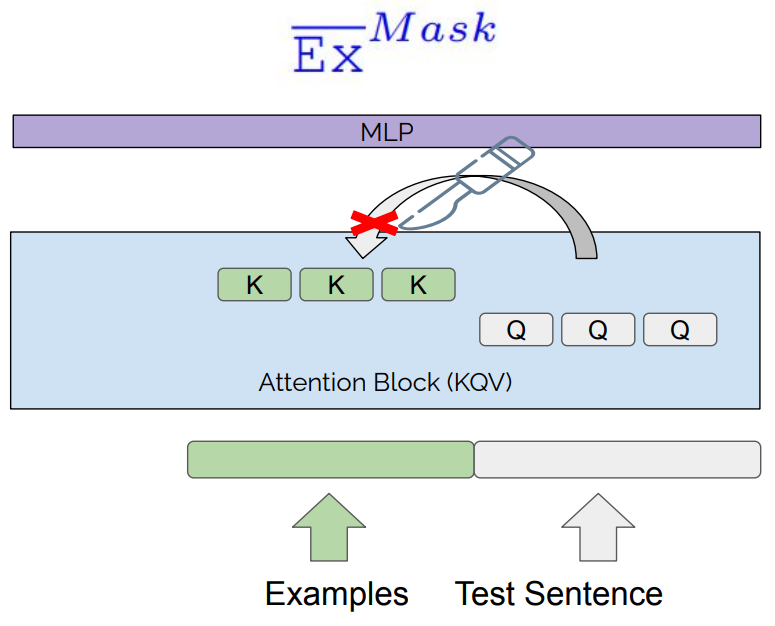}
    \hfill
    \includegraphics[width=0.6\columnwidth,trim=0 0 0 0]{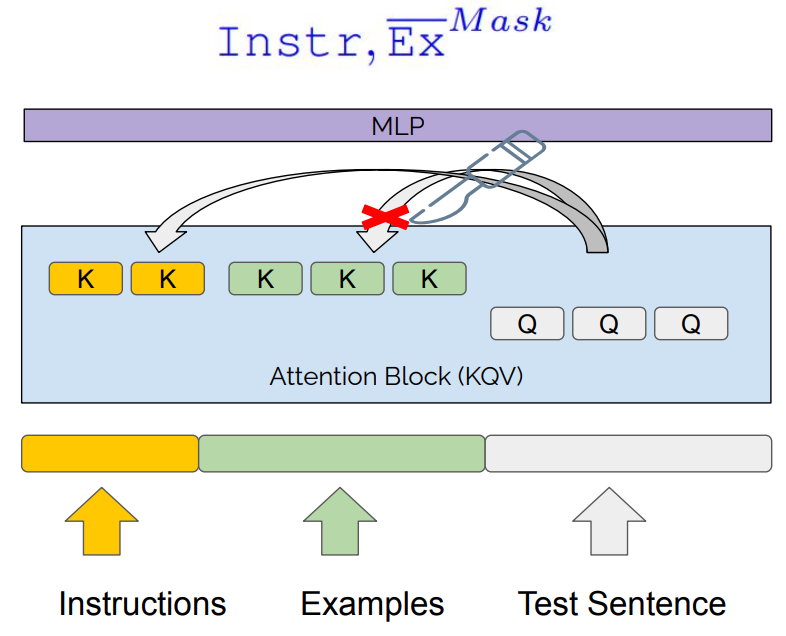}
    \hfill
    \includegraphics[width=0.6\columnwidth,trim=0 0 0 0]{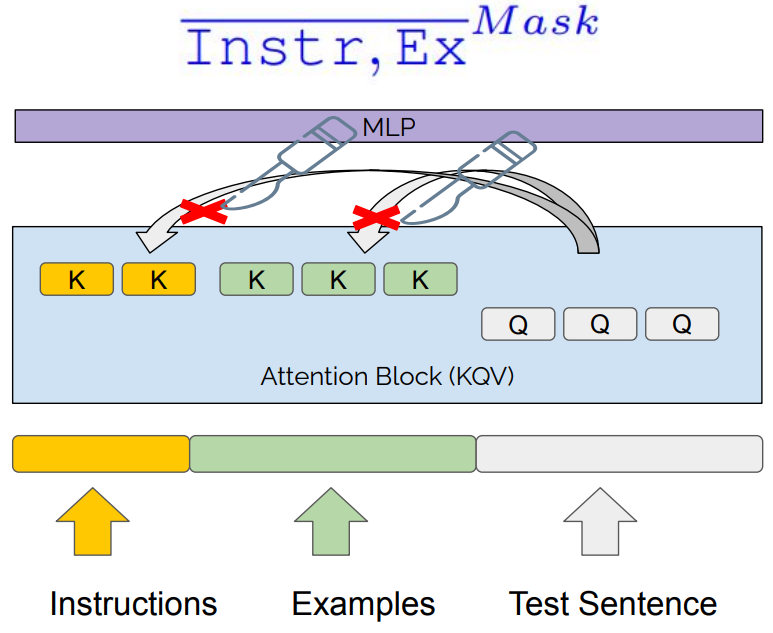}    
    \caption{Graphical explanation of Masking the Attention over Instructions and Examples. The leftmost picture has no instructions and masks examples ($\overline{\texttt{Ex}}^{Mask}$), the middle picture has instructions and masks examples ($\texttt{Instr},\overline{\texttt{Ex}}^{Mask}$), and the rightmost picture masks both instructions and examples ($\overline{\texttt{Instr, Ex}}^{Mask}$).}
    \label{fig:context_mask_graphics}
    
\end{figure*}

\subsection{Prompt Format}
\begin{table*}[!h]
\centering
\begin{small}
\begin{tabular}{lll}
\toprule
Translate English to French. & & \\
Q: A discomfort which lasts .. & A: & Un malaise qui dure
 \\
Q: HTML is a language for formatting & A: & HTML est un langage de formatage \\

... &  & ...\\
Q: After you become comfortable with formatting .. & A: & \\
\bottomrule
\end{tabular}
\end{small} 
\caption{A single continuous input sequence presented to the model for decoding a single test source sentence ``After you become comfortable with formatting..''. Given the entire sequence as input, the model proceeds to generate the target sequence.}
\label{tab:format_example}
\end{table*}



\subsection{Additional Results on English \& Spanish}
\label{sec:appendix-spanish}

In addition to the language pairs $\texttt{en}\rightarrow \texttt{fr}$ and $\texttt{fr}\rightarrow \texttt{en}$, we also run experiments on English and Spanish language pairs, both $\texttt{en}\rightarrow \texttt{es}$ and $\texttt{es}\rightarrow \texttt{en}$.
Due to space limitations, we plot the results of those experiments here. Overall, we see largely identical trends on both directions of English and Spanish to what we observe on English and French translation tasks, leading us to conclude that our conclusions generalize across different translation tasks.

\begin{figure*}[!t]
\centering
 \includegraphics[width=\columnwidth,trim=0 0 0 0]{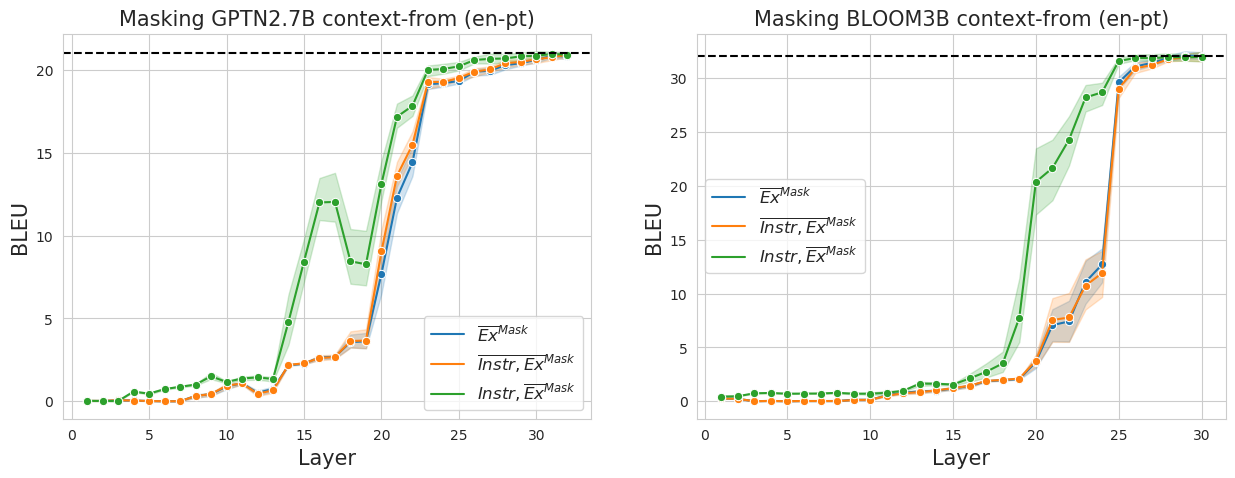}
    \hfill
    \includegraphics[width=\columnwidth,trim=0 0 0 0]{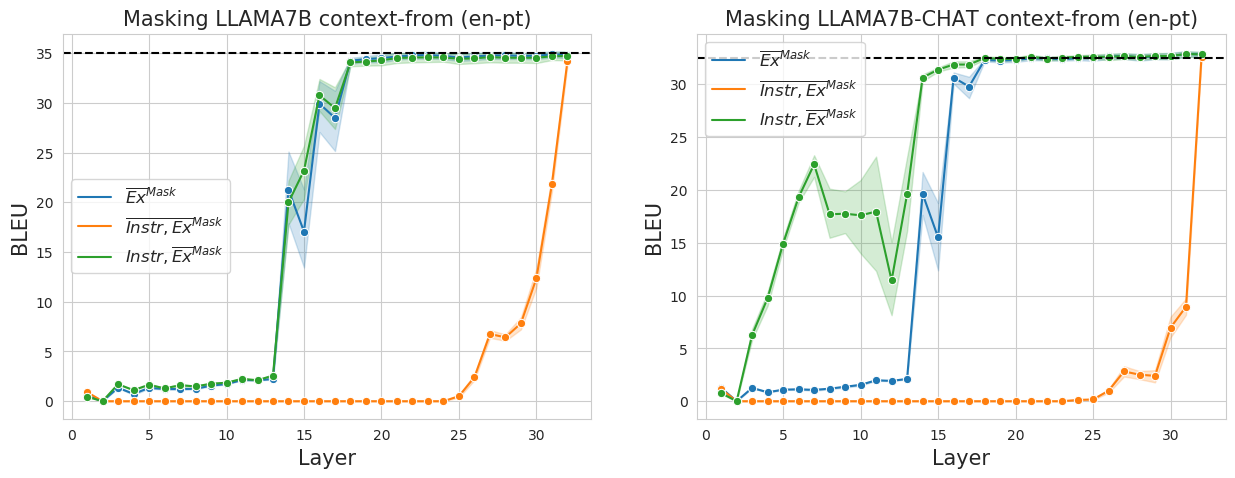}
    \hfill
    \includegraphics[width=\columnwidth,trim=0 0 0 0]{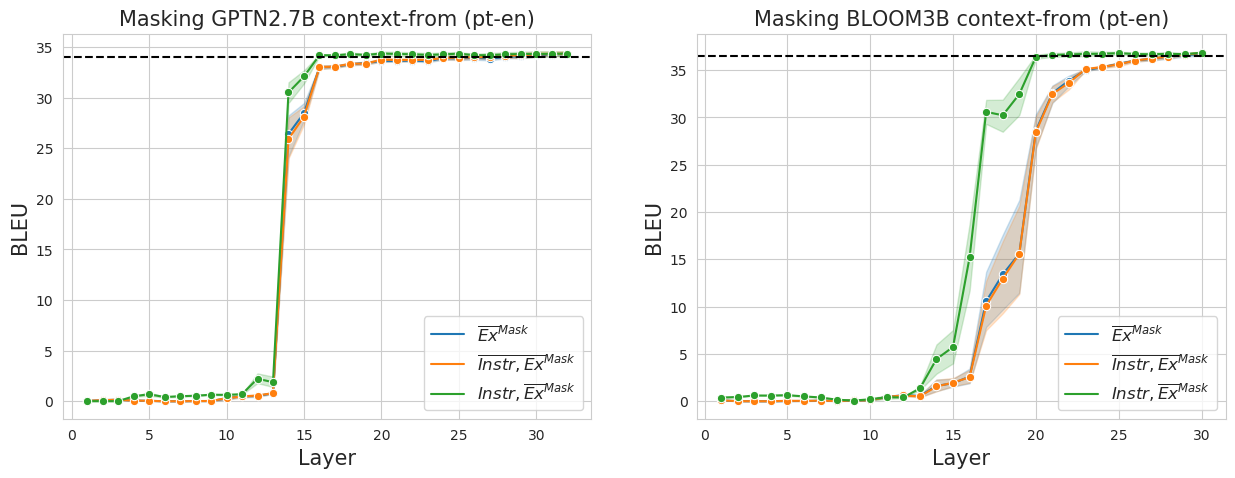}
    \includegraphics[width=\columnwidth,trim=0 0 0 0]{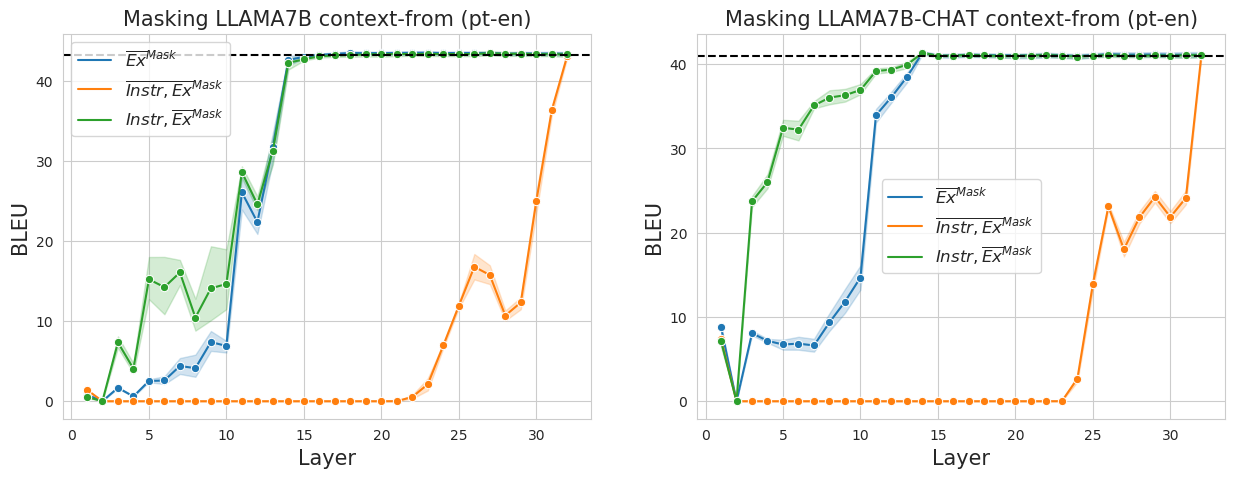}
    \caption{Context-masking and Layer-masking results on the {\bf English $\leftrightarrow$ Portugese} translation task. Critically, we see nearly identical trends to what we see in \autoref{fig:context_mask_fig1} and \autoref{fig:mask_out_layer_gptn} on the English to French translation task, suggesting our results generalize across language pairs.}
    \label{fig:en-es-results}
    
\end{figure*}

\subsection{Autoregressive Decoder only Transformer}

The transformer consists of stacked blocks of self-attention, which itself consists of smaller units of self-attention heads that are concatenated before being fed through a fully connected layer. In autoregressive decoder-only transformers, training and inference adopts a causal mask, where current positions are only able to attend to previous timesteps, instead of being able to attend to the entire input sequence. Unlike encoder-decoder NMT models where source and target sentence have separate processing transformer blocks, decoder-only means that the same model weights are both used to ``encode" the source sentence and ``decode" the target sentence in a single continuous sequence.

\subsection{Training with Autoregressive Translation}
\label{sec:masked_training}

The original language modeling objective in GPT training involves predicting the entire input token sequence which consists of both the source and target sentence (shifted by 1 position). We found this to produce slightly worse results than only minimising the negative log likelihood of predicting the target sentence to be translated, and not the entire sequence. We consider this autoregressive translation training. 


\subsection{$L_0$ Attention Gate Training }

\label{sec:L0_train_details}

\paragraph{Training Details}
For \autoref{sec:L0_train}, We train using Adam Optimizer ($\beta_1=0.9, \beta_2=0.999$) with a batch size of 32, and learning rate of 0.001, early stopping patience of 10 and threshold of 0.01. We initialise attention head gates to be 1 instead of random or 0.5 as this leads to faster convergence. We experiment with two different training settings, the \texttt{0-prompts Train} setting and the \texttt{5-prompts Train} setting. As described in \autoref{sec:masked_training}, we train the model by predicting only the target sentence, conditioned on the context. In the 0-prompt setting, the context consists of the instructions and the source sentence to be translated. In the 5-prompt setting, the context consists of the instructions, 5 prompt examples, and the source sentence to be translated.

In the \texttt{0-prompt} setting, the conditional prefix consists of the instructions and the source sentence to be translated. In the \texttt{5-prompt setting}, the conditional prefix consists of the instruction, 5 source target sentence pairs, and the source sentence to be translated.

\paragraph{Data}

We used the first 10,000 lines of $\texttt{en}\!\rightarrow\! \texttt{fr}$ from WMT06 Europarl \cite{koehn2005europarl} for training.\footnote{Data available from \url{https://www.statmt.org/europarl/}}  To test the generalisability of trained attention head gates, we use a different test domain, FLORES \cite{flores} to reflect the scarcity of in-domain data. We also test an additional language direction $\texttt{en}\!\rightarrow\! \texttt{pt}$ in FLORES to see if training can generalise across languages. 

\paragraph{Training Details}
We train using Adam Optimizer ($\beta_1=0.9, \beta_2=0.999$) with a batch size of 32, and learning rate of 0.001. We use a large early stopping patience of 10 and threshold of 0.01, and train for up to 100 epochs. This is due to the nature of $L_0$ training; we do not expect performance to improve over many iterations and would like the attention gates to keep training as long as there is no large loss in performance. We initialise attention head gates to be 1 instead of random or 0.5 as this leads to much faster convergence and better performance. For the regularisation weight $\lambda$, we search over a hyperparameter set of \{$0.1, 0.01, 0.001, 0.0001$\} and found $0.01$ performs best on the validation set.

\subsection{$L_0$ head masking experiments.}
\label{sec:appendix_L0}
Additional experiments on L0 head masking in the fr$\rightarrow$ en and  es$\rightarrow$ en direction. 

\begin{figure}[!h]
    \centering
\includegraphics[width=0.8\columnwidth]{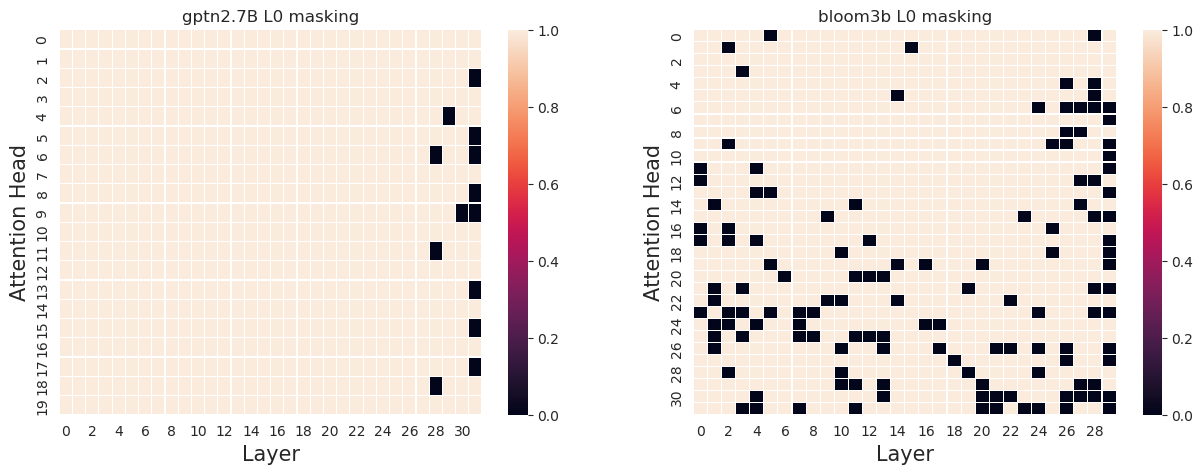}




\includegraphics[width=0.8\columnwidth]{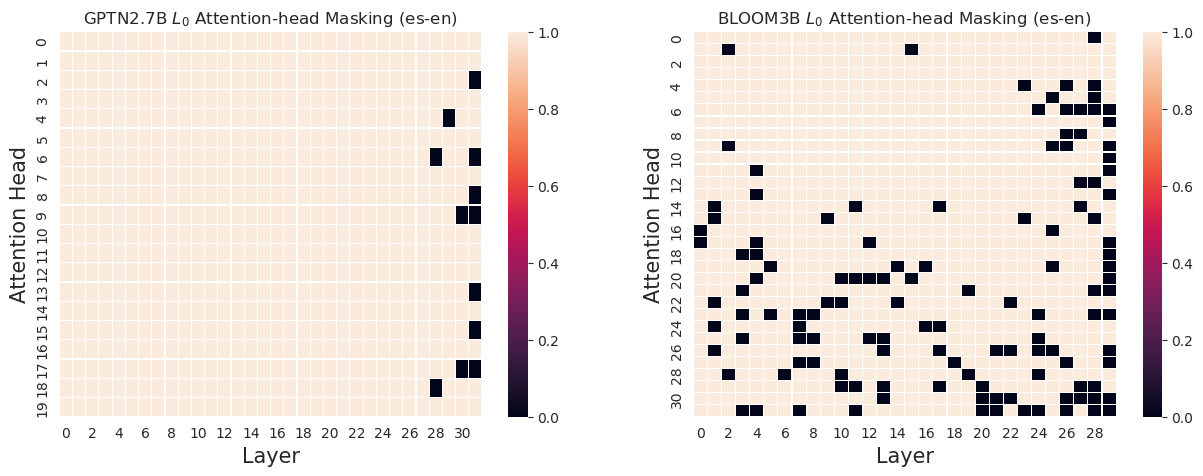}
    \caption{Visualisation of attention head masks for GPTNeo and BLOOM, learned with $L_0 (\lambda=0.01)$ regularisation under a \texttt{0-prompt train} scheme. A value of $0$ (in black) indicates that the attention head is effectively masked out by the trained attention gate. A majority of them occuring at the later layers for GPTNeo and distributed across layers for BLOOM.}
    \label{fig:grid}
\end{figure}


\subsection{Generalisability of $L_0$ gate training}
We experiment with \texttt{0-prompts} and \texttt{5-prompts} in training and using $\lambda\!=\!0$ (no regularisation) and $\lambda\!=\!0.01$. $L_0$ training for the \texttt{0-prompts} shows some gains for the 0-prompts test case, and with no loss on the 5-prompts test case (\autoref{tab:best_ckpt}). Notably, this persists in $\texttt{en}\!\rightarrow \!\texttt{pt}$, a different language direction from training. 

The robustness of translation performance under multiple testing conditions (number of prompts, datasets, language directions) gives some confidence that the trained discrete attention head gates from $L_0$ support a general ability to translate (\autoref{tab:best_ckpt}). In contrast, the soft attention head gates without regularisation ($\lambda=0$) appear to overfit as they perform well on some conditions but deteriorate in others. 

We observe that \texttt{0-prompt} training for $L_0 (\lambda=0.01)$ also outperforms $\texttt{5-prompts}$ which is slightly suprising since \texttt{5-prompts} has more information in the prefix to locate the translation task. One possibility is that the model overfit to the Europarl domain where the training prompts were drawn from. 

\begin{table*}[]
\centering
\resizebox{1.5\columnwidth}{!}{
\begin{tabular}{lr rrrr| lr rrrr}
  & Base & \multicolumn{2}{c}{\texttt{0-prompts} } & \multicolumn{2}{c}{\texttt{5-prompts} } &  & Base & \multicolumn{2}{c}{\texttt{0-prompts} } & \multicolumn{2}{c}{\texttt{5-prompts} } \\
\toprule
 &  & $\lambda\!=\!0$ & $\lambda\!=\!.01$ & $\lambda\!=\!0$ & $\lambda\!=\!.01$ &   &  & $\lambda\!=\!0$ & $\lambda\!=\!.01$ & $\lambda\!=\!0$ & $\lambda\!=\!.01$ \\
\midrule
0-prompts & 18.3 & \colorbox{green}{21.4} & \colorbox{green}{20.1} & \colorbox{green}{18.9} & \colorbox{green}{19.3} & & 6.7 & \colorbox{green}{15.7} & \colorbox{green}{8.6} & \colorbox{green}{13.2} & 6.4 \\
5-prompts & 24.3 & 24.5 & 24.1 & \colorbox{orange}{23.6} & 24.2 & & 25.9 & \colorbox{orange}{19.6} & 25.8 & \colorbox{orange}{24.3} & 26.0 \\
\midrule
\multicolumn{6}{c}{Train: $\texttt{en}\!\rightarrow\! \texttt{fr}$, Test: $\texttt{en}\!\rightarrow\! \texttt{fr}$ } & \multicolumn{6}{c}{Train: $\texttt{en}\!\rightarrow\! \texttt{fr}$, Test: $\texttt{en}\!\rightarrow\! \texttt{pt}$ }
\end{tabular}}
\caption{Performance when using trained attention head gates for $L_0$ with regularisation $\lambda=.01$. $\lambda=0$ refers to training without regularisation. \texttt{0} and \texttt{5} prompts were used in the context for training.  We highlight values which are \colorbox{green}{greater} or \colorbox{orange}{worse} than $0.5$ BLEU points from baseline. Note that as these are compression experiments, we do not expect $L_0$ to perform better than baseline.}
\label{tab:best_ckpt}
\end{table*}


\subsection{Qualitative Analysis of Layer-wise Masking}

\paragraph{\textsc{GPTNeo}} Masking $\ell_{4:8}$ results in a drop in performance for the 0-prompt setting but not the 5-prompt setting (\autoref{fig:mask_out_layer_gptn}), which suggests that $\ell_{4:8}$ are \textbf{not} related to the processing of prompt examples. We emphasise that this interpretation mostly holds at an aggregate level and is not strictly for each instance. For Test Instance ID 575, the model still generates a copy of the English source sentence up to the masking of $\ell_{25}$ for the 0-prompts without instructions setting (\autoref{tab:gen_text_nprompts0_layerwise_noinstr}). This suggests that uncertainty over the task is maintained across layers even though the contributions towards \textit{task location} may be greater from specific layers.               

\paragraph{\textsc{Bloom}} is observed to be more robust to masking of layers; suggesting that task location is more distributed. For the 5-prompt setting, the performance only decreases very slightly. For the 0-prompt setting, we observe that similar to \textsc{GPTNeo}, performance drops when masking out the middle layers. At the aggregate level, \textsc{Bloom} appears to still be translating ($>0$ BLEU) even when layers are masked. However we observe that the drop in performance is due to around $40$ to $50$\% of the test sentences scoring $<5$ BLEU points. There is a clear failure to translate, not simply producing poorer translations.

\label{sec:appendix_generations}
\begin{table*}
\tiny
\begin{tabular}{rrlrl}
\toprule
 layer &   id & lang &   BLEU &                                                                                                                                           text \\
\midrule
     1 &  600 &   cy &   0.00 &                                                                                                                                         uffose \\
     1 &  575 &   ca &   0.00 &                                                                                                                                   B marriages\{ \\
  \midrule
     2 &  600 &   et &   0.00 &                                                                                                                                sses room ( I \\
     2 &  575 &   no &   0.00 &                                                                                                                                            NaN \\
     \midrule
     3 &  600 &   fr &   1.90 &                                                                                            C'est la même chose que l'on a fait avec les virus. \\
     3 &  575 &   fr &  88.44 &                                                           Mais il y a beaucoup de choses sur les oiseaux qui ressemblent encore à un dinosaur. \\
     \midrule

     4 &  600 &   no &   0.00 &                                                                                                                                            NaN \\
     4 &  575 &   no &   0.00 &                                                                                                                                            NaN \\
       \midrule

     5 &  600 &   fr &  78.78 &                                       Hershey et Chase ont utilisé des phages, ou des virus, pour implanter leur propre gène dans un bactérie. \\
     5 &  575 &   fr &  72.98 &                                                         Mais il y a beaucoup de choses sur les oiseaux qui ressemblent toujours à un dinosaur. \\
       \midrule

     6 &  600 &   fr &  78.78 &                                       Hershey et Chase ont utilisé des phages, ou des virus, pour implanter leur propre gène dans un bactérie. \\
     6 &  575 &   fr &  60.29 &                                                       Mais il y a beaucoup de choses à propos de oiseaux qui ressemblent encore à un dinosaur. \\
       \midrule

     7 &  600 &   fr &  13.94 &                                                                               Hershey et Chase ont implanté leur propre gène dans un bactérie. \\
     7 &  575 &   fr &  72.98 &                                                         Mais il y a beaucoup de choses sur les oiseaux qui ressemblent toujours à un dinosaur. \\
       \midrule

     8 &  600 &   no &   0.00 &                                                                                                                                            NaN \\
     8 &  575 &   fr &  88.44 &                                                            Mais il y a beaucoup de choses sur les oiseaux qui ressemblent encore à un dinosau. \\
      \midrule

     9 &  600 &   no &   0.00 &                                                                                                                                            NaN \\
     9 &  575 &   fr &  82.82 &                                                        Mais il y a beaucoup de choses sur les oiseaux qui ressemblent toujours à un dinosaure. \\
     \midrule

    10 &  600 &   no &   0.00 &                                                                                                                                            NaN \\
    10 &  575 &   en &   2.73 &                                                                     But there are a lot of things about birds that still look like a dinosaur. \\
      \midrule

    11 &  600 &   en &   4.78 &                                                          Hershey and Chase used phages, or viruses, to implant their own DNA into a bacterium. \\
    11 &  575 &   en &   2.73 &                                                                     But there are a lot of things about birds that still look like a dinosaur. \\
      \midrule

    12 &  600 &   no &   0.00 &                                                                                                                                            NaN \\
    12 &  575 &   no &   0.00 &                                                                                                                                            NaN \\
      \midrule

    13 &  600 &   no &   0.00 &                                                                                                                                            NaN \\
    13 &  575 &   fr &  35.75 &                                                                                        Mais il y a beaucoup de choses que je ne comprends pas. \\
     \midrule

    14 &  600 &   en &   4.78 &                                                          Hershey and Chase used phages, or viruses, to implant their own DNA into a bacterium. \\
    14 &  575 &   en &   2.73 &                                                                     But there are a lot of things about birds that still look like a dinosaur. \\
     \midrule

    15 &  600 &   fr &  76.48 &                                       Hershey et Chase ont utilisé des phages, ou des virus, pour implanter leur propre gène dans un bacillus. \\
    15 &  575 &   en &   2.73 &                                                                     But there are a lot of things about birds that still look like a dinosaur. \\
    \midrule

    16 &  600 &   fr &  78.78 &                                       Hershey et Chase ont utilisé des phages, ou des virus, pour implanter leur propre gène dans un bactérie. \\
    16 &  575 &   fr &  70.18 &                                                   Mais il y a beaucoup de choses sur les oiseaux qui ressemblent toujours comme un dinosaurof. \\
     \midrule

    17 &  600 &   fr &  82.32 &                                  Les Hershey et Chase ont utilisé des phages, ou des virus, pour implanter leur propre gène dans une bactérie. \\
    17 &  575 &   fr &  88.44 &                                                           Mais il y a beaucoup de choses sur les oiseaux qui ressemblent encore à un dinosaur. \\
      \midrule

    18 &  600 &   fr &  78.78 &                                     Hershey et Chase ont utilisé des phages, ou des virus, pour implanter leur propre génome dans un bactérie. \\
    18 &  575 &   fr &  66.38 &                                                            Mais il y a beaucoup de choses sur les oiseaux qui aussi ressemble à un dinosaures. \\
     \midrule

    19 &  600 &   fr &  59.33 &                   Les héritiers de Hershey et de Chase ont utilisé des phages, ou des virus, pour implanter leur propre gène dans un bactérie. \\
    19 &  575 &   fr &  47.91 &                                                 Mais il y a beaucoup de choses à propos de les oiseaux qui ressemblent toujours à un dinosaur. \\
     \midrule

    20 &  600 &   fr &  48.82 &                                       Hershey et Chase ont utilisé les phages, ou les virus, pour implanter leur propre gène dans un bactérie. \\
    20 &  575 &   en &   2.73 &                                                                     But there are a lot of things about birds that still look like a dinosaur. \\
     \midrule

    21 &  600 &   fr &  78.78 &                                       Hershey et Chase ont utilisé des phages, ou des virus, pour implanter leur propre gène dans un bactérie. \\
    21 &  575 &   fr &  88.44 &                                                           Mais il y a beaucoup de choses sur les oiseaux qui ressemblent encore à un dinosaur. \\
      \midrule

    22 &  600 &   fr &  48.82 &                                       Hershey et Chase ont utilisé les phages, ou les virus, pour implanter leur propre gène dans un bactérie. \\
    22 &  575 &   fr &  88.44 &                                                         Mais il y a beaucoup de choses sur les oiseaux qui ressemblent encore à un dinosaurof. \\
      \midrule

    23 &  600 &   fr &  78.78 &                                       Hershey et Chase ont utilisé des phages, ou des virus, pour implanter leur propre gène dans un bactérie. \\
    23 &  575 &   fr &  88.44 &                                                           Mais il y a beaucoup de choses sur les oiseaux qui ressemblent encore à un dinosaur. \\
      \midrule

    24 &  600 &   fr &  78.78 &                                     Hershey et Chase ont utilisé des phages, ou des virus, pour implanter leur propre génome dans un bactérie. \\
    24 &  575 &   fr &  62.72 &                                                   Mais il y a beaucoup de choses à propos de les oiseaux qui ressemblent encore à un dinosaur. \\
      \midrule

    25 &  600 &   fr &  78.78 &                                       Hershey et Chase ont utilisé des phages, ou des virus, pour implanter leur propre gène dans un bactérie. \\
    25 &  575 &   fr &  88.44 &                                                           Mais il y a beaucoup de choses sur les oiseaux qui ressemblent encore à un dinosaur. \\
      \midrule

    26 &  600 &   fr &  78.78 &                                       Hershey et Chase ont utilisé des phages, ou des virus, pour implanter leur propre gène dans un bactérie. \\
    26 &  575 &   fr &  88.44 &                                                           Mais il y a beaucoup de choses sur les oiseaux qui ressemblent encore à un dinosaur. \\
      \midrule

    27 &  600 &   fr &  66.28 &                                     Hershey et Château ont utilisé des phages, ou des virus, pour implanter leur propre gène dans un bactérie. \\
    27 &  575 &   fr &  88.44 &                                                           Mais il y a beaucoup de choses sur les oiseaux qui ressemblent encore à un dinosaur. \\
      \midrule

    28 &  600 &   fr &  78.78 &                                       Hershey et Chase ont utilisé des phages, ou des virus, pour implanter leur propre gène dans un bactérie. \\
    28 &  575 &   fr &  88.44 &                                                           Mais il y a beaucoup de choses sur les oiseaux qui ressemblent encore à un dinosaur. \\
      \midrule

    29 &  600 &   fr &  59.33 &                   Les héritiers de Hershey et de Chase ont utilisé des phages, ou des virus, pour implanter leur propre gène dans un bactérie. \\
    29 &  575 &   fr &  88.44 &                                                         Mais il y a beaucoup de choses sur les oiseaux qui ressemblent encore à un dinosaurof. \\
      \midrule

    30 &  600 &   fr &  78.78 &                                       Hershey et Chase ont utilisé des phages, ou des virus, pour implanter leur propre gène dans un bactérie. \\
    30 &  575 &   fr &  88.44 &                                                           Mais il y a beaucoup de choses sur les oiseaux qui ressemblent encore à un dinosaur. \\
      \midrule

    31 &  600 &   fr &  78.78 &                                       Hershey et Chase ont utilisé des phages, ou des virus, pour implanter leur propre gène dans un bactérie. \\
    31 &  575 &   fr &  88.44 &                                                           Mais il y a beaucoup de choses sur les oiseaux qui ressemblent encore à un dinosaur. \\
      \midrule

    32 &  600 &   fr &   6.44 &                                                         Les héritiers de Hershey et de Chase ont été capables de l'implanter dans un bactérie. \\
    32 &  575 &   fr &  51.52 &                                                  Mais il y a beaucoup de choses sur les oiseaux que sont encore aussi vus comme un dinosaures. \\
\bottomrule
\end{tabular}
\caption{0-prompts with instructions, masking layer by layer of \textsc{GPTNeo}2.7B}
\label{tab:gen_text_nprompts0_layerwise}
\end{table*}

\begin{table*}
\tiny
\begin{tabular}{rrlrl}
\toprule
 layer &   id & lang &  BLEU &                              text \\
\midrule
     1 &  575 &   no &   0.0 &                                                                                                                                                                                                                             NaN \\
     1 &  600 &   no &   0.0 &                                                                                                                                                                                                                             NaN \\
     \midrule
     2 &  575 &  NaN &   0.0 &                                                                                                                                                                                                                               , \\
     2 &  600 &   no &   0.0 &                                                                                                                                                                                                                             NaN \\
     \midrule
     3 &  575 &   fr &   2.3 &                                                                                                                                                                                         C'est pas un oiseau, c'est un dinosaur. \\
     3 &  600 &   fr &   0.8 &                                   [phare] Phare, phare, phare, phare, phare, phare, phare, phare, phare, phare, phare, ..., \\
     \midrule
     4 &  575 &   en &   2.7 &                                                                                                                                                                             "I think it's a dinosaur, I think it's a dinosaur." \\
     4 &  600 &   fr &   2.6 &                                                                               Les virus, c'est-ce qu'on dit? C'un mot? C'est pas un mot? C'un mot? C'un'un? ... \\
     \midrule
     5 &  575 &   fr &  42.9 &                                                                                                                                  Mais il y a beaucoup de choses à propos de oiseaux qui ressemblent toujours comme un dinosaur. \\
     5 &  600 &   fr &  73.6 &                                                                                                                      L'Hershey et Chase ont utilisé des phages, ou des virus, pour implanter leur propre gène dans un bactérie. \\
     \midrule
     6 &  575 &   fr &  53.8 &                                                                                                                                     Mais il y a beaucoup de choses à propos de oiseaux qui ressemblent toujours à un dinosaure. \\
     6 &  600 &   fr &  74.9 &                                                                                                                    Les Hershey et Chase ont utilisé des phages, ou des virus, pour implanter leur propre gène dans un bactérie. \\
     \midrule
     7 &  575 &   fr &  76.2 &                                                                                                                                           Et il y a beaucoup de choses sur les oiseaux qui ressemblent toujours à un dinosaure. \\
     7 &  600 &   fr &  83.8 &                                                                                                                     Les Hershey et Chase ont utilisé des phages, ou des virus, pour implanter leur propre ADN dans un bactérie. \\
     \midrule
     8 &  575 &   fr &  88.4 &                                                                                                                                             Mais il y a beaucoup de choses sur les oiseaux qui ressemblent encore à un dinosau. \\
     8 &  600 &   fr &  83.8 &                                                                                                                 L'usine Hershey et Chase ont utilisé des phages, ou des virus, pour implanter leur propre ADN dans un bactérie. \\
     \midrule
     9 &  575 &   en &   1.5 &                                                                                                                                                                                                         The bird is a dinosaur. \\
     9 &  600 &   fr &  33.9 &                                                                                                                  Les hémoglobine et Chase utilisent les phages, ou les virus, pour implanter leur propre gène dans un bactérie. \\
    \midrule
    10 &  575 &   en &   2.7 &                                                                                                                                                      But there are a lot of things about birds that still look like a dinosaur. \\
    10 &  600 &   fr &  11.4 &                                                                                                                                                           Les phages, ou virus, ont implanté leur propre gène dans un bactérie. \\
    \midrule
    11 &  575 &   en &   2.6 &                                                                                                                                                         I think it's a good idea to have a little bit of a bird in your pocket. \\
    11 &  600 &   en &   0.0 &  The French have a saying: "The French have a saying: "The French have a saying: "The French have a saying:... \\
    \midrule
    12 &  575 &   en &   2.7 &                                                                                                                                                      But there are a lot of things about birds that still look like a dinosaur. \\
    12 &  600 &   en &   1.7 &                                                                                                                                                          The bacterium was then able to use the phage to infect other bacteria. \\
    \midrule
    13 &  575 &   en &   2.7 &                                                                                                                                                      But there are a lot of things about birds that still look like a dinosaur. \\
    13 &  600 &   fr &  18.7 &                                                                                                                                    L'entreprise Hershey a utilisé des phages pour implanter leur propre DNA dans leur bactérie. \\
    \midrule
    14 &  575 &   en &   2.7 &                                                                                                                                                      But there are a lot of things about birds that still look like a dinosaur. \\
    14 &  600 &   en &   4.8 &                                                                                                                                           Hershey and Chase used phages, or viruses, to implant their own DNA into a bacterium. \\
    \midrule
    15 &  575 &   fr &   3.0 &                                                                                                                                                                           C'est pas un truc de poulet, c'est un truc de poulet. \\
    15 &  600 &   fr &  35.7 &                                                                                                                    L'université de Paris-Sud a utilisé des phages, ou viraux, pour implanter leur propre gène dans un bacillus. \\
    \midrule
    16 &  575 &   en &   2.7 &                                                                                                                                                      But there are a lot of things about birds that still look like a dinosaur. \\
    16 &  600 &   fr &  74.9 &                                                                                                                  Les Hershey et Chase ont utilisé des phages, ou des virus, pour implanter leur propre génome dans un bactérie. \\
    \midrule
    17 &  575 &   en &   2.7 &                                                                                                                                                      But there are a lot of things about birds that still look like a dinosaur. \\
    17 &  600 &   fr &  82.3 &                                                                                                                   Les Hershey et Chase ont utilisé des phages, ou des virus, pour implanter leur propre gène dans une bactérie. \\
    \midrule
    18 &  575 &   en &   2.7 &                                                                                                                                                      But there are a lot of things about birds that still look like a dinosaur. \\
    18 &  600 &   fr &  74.9 &                                                                                                                  Les Hershey et Chase ont utilisé des phages, ou des virus, pour implanter leur propre génome dans un bactérie. \\
    \midrule
    19 &  575 &   fr &  44.2 &                                                                                                                                      Mais il y a beaucoup de choses à propos de oiseaux qui ressemblent toujours à un dinosaur. \\
    19 &  600 &   fr &  59.3 &                                                                                                    Les héritiers de Hershey et de Chase ont utilisé des phages, ou des virus, pour implanter leur propre gène dans un bactérie. \\
    \midrule
    20 &  575 &   en &   2.7 &                                                                                                                                                      But there are a lot of things about birds that still look like a dinosaur. \\
    20 &  600 &   fr &  46.4 &                                                                                                                    Les Hershey et Chase ont utilisé les phages, ou les virus, pour implanter leur propre gène dans un bactérie. \\
    \midrule
    21 &  575 &   en &   2.7 &                                                                                                                                                      But there are a lot of things about birds that still look like a dinosaur. \\
    21 &  600 &   fr &  74.9 &                                                                                                                L'usine Hershey et Chase ont utilisé des phages, ou des virus, pour implanter leur propre gène dans un bactérie. \\
    \midrule
    22 &  575 &   en &   2.7 &                                                                                                                                                      But there are a lot of things about birds that still look like a dinosaur. \\
    22 &  600 &   fr &  56.3 &                                                                                                                     Les Hershey et Chase ont utilisé les phages, ou les virus, pour implanter leur propre ADN dans un bactérie. \\
    \midrule
    23 &  575 &   en &   2.7 &                                                                                                                                                      But there are a lot of things about birds that still look like a dinosaur. \\
    23 &  600 &   fr &  82.9 &                                                                                                                       L'Hershey et Chase ont utilisé des phages, ou des virus, pour implanter leur propre ADN dans un bactérie. \\
    \midrule
    24 &  575 &   fr &  60.3 &                                                                                                                                        Mais il y a beaucoup de choses à propos de oiseaux qui ressemblent encore à un dinosaur. \\
    24 &  600 &   fr &  37.0 &                                                                                                              L'usine Hershey et Chase utilisaient les phages, ou les virus, pour implanter leur propre génome dans un bactérie. \\
    \midrule
    25 &  575 &   en &   2.7 &                                                                                                                                                      But there are a lot of things about birds that still look like a dinosaur. \\
    25 &  600 &   fr &  74.9 &                                                                                                                    Les Hershey et Chase ont utilisé des phages, ou des virus, pour implanter leur propre gène dans un bactérie. \\
    \midrule
    26 &  575 &   fr &  71.6 &                                                                                                                                       Mais il y a beaucoup de choses à propos de oiseaux qui ressemblent encore à un dinosaure. \\
    26 &  600 &   fr &  73.6 &                                                                                                                      L'Hershey et Chase ont utilisé des phages, ou des virus, pour implanter leur propre gène dans un bactérie. \\
    \midrule
    27 &  575 &   fr &  88.4 &                                                                                                                                          Mais il y a beaucoup de choses sur les oiseaux qui ressemblent encore à un dinosaurof. \\
    27 &  600 &   fr &  63.0 &                                                                                                                  Les Hershey et Château ont utilisé des phages, ou des virus, pour implanter leur propre gène dans un bactérie. \\
    \midrule
    28 &  575 &   fr &  44.2 &                                                                                                                                      Mais il y a beaucoup de choses à propos de oiseaux qui ressemblent toujours à un dinosaur. \\
    28 &  600 &   fr &  74.9 &                                                                                                                    Les Hershey et Chase ont utilisé des phages, ou des virus, pour implanter leur propre gène dans un bactérie. \\
    \midrule
    29 &  575 &   fr &  87.0 &                                                                                                                                     Mais il y a beaucoup de choses sur les oiseaux qui ressemblent encore à un dinosaurofrench: \\
    29 &  600 &   fr &  53.4 &                                                                                                        L'entreprise de la filière Hershey a utilisé des phages, ou des virus, pour implanter leur propre gène dans un bactérie. \\
    \midrule
    30 &  575 &   fr &  82.8 &                                                                                                                                         Mais il y a beaucoup de choses sur les oiseaux qui ressemblent toujours à un dinosaure. \\
    30 &  600 &   fr &  74.9 &                                                                                                                    Les Hershey et Chase ont utilisé des phages, ou des virus, pour implanter leur propre gène dans un bactérie. \\
    \midrule
    31 &  575 &   fr &  82.8 &                                                                                                                                         Mais il y a beaucoup de choses sur les oiseaux qui ressemblent toujours à un dinosaure. \\
    31 &  600 &   fr &  59.3 &                                                                                                  Les hémoglobins de Hershey et de Chase ont utilisé des phages, ou des virus, pour implanter leur propre gène dans un bactérie. \\
    \midrule
    32 &  575 &   fr &  67.5 &                                                                                                                                           Mais il y a beaucoup de choses sur les oiseaux qui aussi ressemblent à un dinosaurof. \\
    32 &  600 &   fr &   6.7 &                                                                                                                                        L'Hershey et le Chase ont été capables de l'implanter leur propre gène dans un bactérie. \\
\bottomrule
\end{tabular}
\caption{0-prompts without instructions, masking layer by layer of \textsc{GPTNeo}2.7B}
\label{tab:gen_text_nprompts0_layerwise_noinstr}
\end{table*}
\begin{table*}
\tiny
\begin{tabular}{rrlrl}
\toprule
 layer &   id & lang &   BLEU &                                                                                                                                                                                                                       text \\
\midrule
     1 &  902 &   en &    0.0 &                                                                                                                                                                                                                       : of \\
     1 &  575 &   en &    0.0 &                                                                                                                                                                                                                         of \\
     \midrule
     2 &  902 &   en &    0.0 &                                                                                                                                                                                           of(n, very very-   ofS First \\
     2 &  575 &   da &    0.0 &                                                                                                                                                                                                                         f( \\
     \midrule
     3 &  902 &   fr &  100.0 &                                                                                                                                   Les scènes sont affichées sur les pyramides et les différentes pyramides sont éclairées. \\
     3 &  575 &   fr &   88.4 &                                                                                                                                       Mais il y a beaucoup de choses sur les oiseaux qui ressemblent encore à un dinosaur. \\
     \midrule
     4 &  902 &   en &    0.0 &                                       the, the French, the French, the English, the, the, the, the, the, the, the, the, the, the,  \\
     4 &  575 &   no &    0.0 &                                                    NaN \\
     \midrule
     5 &  902 &   fr &   65.9 &                                                                                                                                     Les scènes sont affichées sur les pyramides et les différents pyramides sont éclairés. \\
     5 &  575 &   fr &   88.4 &                                                                                                                                       Mais il y a beaucoup de choses sur les oiseaux qui ressemblent encore à un dinosaur. \\
     \midrule
     6 &  902 &   fr &  100.0 &                                                                                                                                   Les scènes sont affichées sur les pyramides et les différentes pyramides sont éclairées. \\
     6 &  575 &   fr &   88.4 &                                                                                                                                       Mais il y a beaucoup de choses sur les oiseaux qui ressemblent encore à un dinosaur. \\
     \midrule
     7 &  902 &   fr &  100.0 &                                                                                                                                   Les scènes sont affichées sur les pyramides et les différentes pyramides sont éclairées. \\
     7 &  575 &   fr &  100.0 &                                                                                                                                      Mais il y a beaucoup de choses sur les oiseaux qui ressemblent encore à un dinosaure. \\
     \midrule
     8 &  902 &   fr &  100.0 &                                                                                                                                   Les scènes sont affichées sur les pyramides et les différentes pyramides sont éclairées. \\
     8 &  575 &   fr &  100.0 &                                                                                                                                      Mais il y a beaucoup de choses sur les oiseaux qui ressemblent encore à un dinosaure. \\
     \midrule
     9 &  902 &   fr &   65.9 &                                                                                                                                     Les scènes sont affichées sur les pyramides et les différents pyramides sont éclairés. \\
     9 &  575 &   fr &  100.0 &                                                                                                                                      Mais il y a beaucoup de choses sur les oiseaux qui ressemblent encore à un dinosaure. \\
    \midrule
    10 &  902 &   fr &  100.0 &                                                                                                                                   Les scènes sont affichées sur les pyramides et les différentes pyramides sont éclairées. \\
    10 &  575 &   fr &  100.0 &                                                                                                                                      Mais il y a beaucoup de choses sur les oiseaux qui ressemblent encore à un dinosaure. \\
    \midrule
    11 &  902 &   fr &    1.4 &   Fondamentalement, vous afficherez des annonces pour proposer votre aide, arpenterez les quais, ... \\
    11 &  575 &   fr &    1.3 &  Fondamentalement, vous afficherez des annonces pour proposer votre aide, arpenterez les quais, ... \\
    \midrule
    12 &  902 &   fr &  100.0 &                                                                                                                                   Les scènes sont affichées sur les pyramides et les différentes pyramides sont éclairées. \\
    12 &  575 &   fr &   42.5 &                                                                                                                                                              Mais il y a beaucoup de choses qui ressemblent à un dinosaur. \\
    \midrule
    13 &  902 &   fr &   34.5 &                                                                                                                             Les scènes sont déclarées sur les pyramides et les pyramides sont déclarées sur les pyramides. \\
    13 &  575 &   fr &    5.5 &                                                                                                                                                       Les oiseaux sont des animaux, mais ils sont aussi des êtres humains. \\
    \midrule
    14 &  902 &   fr &  100.0 &                                                                                                                                   Les scènes sont affichées sur les pyramides et les différentes pyramides sont éclairées. \\
    14 &  575 &   fr &   73.3 &                                                                                                                                         Mais il y a beaucoup de choses sur les oiseaux qui ressemblent à un dinosauresque. \\
    \midrule
    15 &  902 &   fr &   64.0 &                                                                                                                                     Les scènes sont affichées sur les pyramides et les pyramides différents sont éclairés. \\
    15 &  575 &   fr &   26.7 &                                                                                                       Mais il y a beaucoup de choses à propos de la façon dont les oiseaux se ressemblent, même si c'est un peu plus tard. \\
    \midrule
    16 &  902 &   fr &  100.0 &                                                                                                                                   Les scènes sont affichées sur les pyramides et les différentes pyramides sont éclairées. \\
    16 &  575 &   fr &   76.7 &                                                                                                                                 Mais il y a beaucoup de choses sur les oiseaux qui ressemblent encore comme un dinosaures. \\
    \midrule
    17 &  902 &   fr &  100.0 &                                                                                                                                   Les scènes sont affichées sur les pyramides et les différentes pyramides sont éclairées. \\
    17 &  575 &   fr &   88.4 &                                                                                                                                       Mais il y a beaucoup de choses sur les oiseaux qui ressemblent encore à un dinosaur. \\
    \midrule
    18 &  902 &   fr &   65.9 &                                                                                                                                     Les scènes sont affichées sur les pyramides et les différents pyramides sont éclairés. \\
    18 &  575 &   fr &   88.4 &                                                                                                                                  Mais il y a beaucoup de choses sur les oiseaux qui ressemblent encore à un dinosauresque. \\
    \midrule
    19 &  902 &   fr &   65.9 &                                                                                                                                    Les scènes sont affichées sur les pyramides et les différents pyramides sont illuminés. \\
    19 &  575 &   fr &   88.4 &                                                                                                                                       Mais il y a beaucoup de choses sur les oiseaux qui ressemblent encore à un dinosaur. \\
    \midrule
    20 &  902 &   fr &   65.9 &                                                                                                                                     Les scènes sont affichées sur les pyramides et les différents pyramides sont éclairés. \\
    
    20 &  575 &   fr &   88.4 &                                                                                                                                       Mais il y a beaucoup de choses sur les oiseaux qui ressemblent encore à un dinosaur. \\
    \midrule
    21 &  902 &   fr &  100.0 &                                                                                                                                   Les scènes sont affichées sur les pyramides et les différentes pyramides sont éclairées. \\
    21 &  575 &   fr &   88.4 &                                                                                                                                       Mais il y a beaucoup de choses sur les oiseaux qui ressemblent encore à un dinosaur. \\
    \midrule
    22 &  902 &   fr &   64.0 &                                                                                                                                     Les scènes sont affichées sur les pyramides et les pyramides différents sont éclairés. \\
    22 &  575 &   fr &   88.4 &                                                                                                                                   Mais il y a beaucoup de choses sur les oiseaux qui ressemblent encore à un dinosauroide. \\
    \midrule
    23 &  902 &   fr &   65.9 &                                                                                                                                     Les scènes sont affichées sur les pyramides et les différents pyramides sont éclairés. \\
    23 &  575 &   fr &   88.4 &                                                                                                                                       Mais il y a beaucoup de choses sur les oiseaux qui ressemblent encore à un dinosaur. \\
    \midrule
    24 &  902 &   fr &   78.3 &                                                                                                                                    Les scènes sont affichées sur les pyramides et les différents pyramides sont éclairées. \\
    24 &  575 &   fr &   88.4 &                                                                                                                                       Mais il y a beaucoup de choses sur les oiseaux qui ressemblent encore à un dinosaur. \\
    \midrule
    25 &  902 &   fr &  100.0 &                                                                                                                                   Les scènes sont affichées sur les pyramides et les différentes pyramides sont éclairées. \\
    25 &  575 &   fr &   88.4 &                                                                                                                                       Mais il y a beaucoup de choses sur les oiseaux qui ressemblent encore à un dinosaur. \\
    \midrule
    26 &  902 &   fr &   65.9 &                                                                                                                                     Les scènes sont affichées sur les pyramides et les différents pyramides sont éclairés. \\
    26 &  575 &   fr &   88.4 &                                                                                                                                       Mais il y a beaucoup de choses sur les oiseaux qui ressemblent encore à un dinosaur. \\
    \midrule
    27 &  902 &   fr &  100.0 &                                                                                                                                   Les scènes sont affichées sur les pyramides et les différentes pyramides sont éclairées. \\
    27 &  575 &   fr &   88.4 &                                                                                                                                       Mais il y a beaucoup de choses sur les oiseaux qui ressemblent encore à un dinosaur. \\
    \midrule
    28 &  902 &   fr &   65.9 &                                                                                                                                     Les scènes sont affichées sur les pyramides et les différents pyramides sont éclairés. \\
    28 &  575 &   fr &   88.4 &                                                                                                                                       Mais il y a beaucoup de choses sur les oiseaux qui ressemblent encore à un dinosaur. \\
    \midrule
    29 &  902 &   fr &   65.9 &                                                                                                                                     Les scènes sont affichées sur les pyramides et les différents pyramides sont éclairés. \\
    29 &  575 &   fr &  100.0 &                                                                                                                                      Mais il y a beaucoup de choses sur les oiseaux qui ressemblent encore à un dinosaure. \\
    \midrule
    30 &  902 &   fr &   78.3 &                                                                                                                                    Les scènes sont affichées sur les pyramides et les différents pyramides sont éclairées. \\
    30 &  575 &   fr &   88.4 &                                                                                                                                       Mais il y a beaucoup de choses sur les oiseaux qui ressemblent encore à un dinosaur. \\
    \midrule
    31 &  902 &   fr &   65.9 &                                                                                                                                     Les scènes sont affichées sur les pyramides et les différents pyramides sont éclairés. \\
    31 &  575 &   fr &   88.4 &                                                                                                                                       Mais il y a beaucoup de choses sur les oiseaux qui ressemblent encore à un dinosaur. \\
    \midrule
    32 &  902 &   fr &   65.9 &                                                                                                                                   Les scènes sont affichées sur les pyramides et les différents pyramides sont illuminées. \\
    32 &  575 &   fr &   67.5 &                                                                                                                                 Mais il y a beaucoup de choses sur les oiseaux qui aujourd’hui ressemblent à un dinosauro. \\
\bottomrule
\end{tabular}
\caption{5-prompts with instructions, masking layer by layer of \textsc{GPTNeo}2.7B}
\label{tab:gen_text_nprompts5_layerwise}
\end{table*}

\begin{table*}
\tiny
\begin{tabular}{rrlrl}
\toprule
 layer &   id & lang &   BLEU &                                              text \\
\midrule
     1 &  600 &   tl &    0.0 &                                                                                                                                                                                                                                                                          *- ing \\
     1 &  575 &   en &    0.6 &                                                                                                                                                                                                                                                                             fl.of, \\
     \midrule
     2 &  600 &   en &    1.6 &                                                                                                                                                                                         in      " " -   ( --, -,   -  (es,"  " " so   " whats " whats"  " between whats --what  e, \\
     2 &  575 &   en &    2.3 &                                                                                                                                                                                                                                     "    ",what awaited      ico    " " " "\_,    . \\
     \midrule
     3 &  600 &   fr &   86.6 &                                                                                                                                                                          Hershey et Chase ont utilisé des phages, ou des virus, pour implanter leur propre gène dans une bactérie. \\
     3 &  575 &   fr &   88.4 &                                                                                                                                                                                               Mais il y a beaucoup de choses sur les oiseaux qui ressemblent encore à un dinosaur. \\
     \midrule
     4 &  600 &   en &    0.3 &                                                                                                                                                                                                                                                                          the, etc. \\
     4 &  575 &   ro &    0.0 &                                                                                                               are: are: are: are: are: are: are: are: are: are: are: are: are: are: are: are: are: ... \\
     \midrule
     5 &  600 &   fr &   76.5 &                                                                                                                                                                            Hershey et Chase ont utilisé des phages, ou des virus, pour implanter leur propre gène dans un bacille. \\
     5 &  575 &   fr &   88.4 &                                                                                                                                                                                               Mais il y a beaucoup de choses sur les oiseaux qui ressemblent encore à un dinosaur. \\
     \midrule
     6 &  600 &   fr &   88.1 &                                                                                                                                                                            Hershey et Chase ont utilisé des phages, ou des virus, pour implanter leur propre ADN dans un bactérie. \\
     6 &  575 &   fr &   62.7 &                                                                                                                                                                                       Mais il y a beaucoup de choses à propos de les oiseaux qui ressemblent encore à un dinosaur. \\
     \midrule
     7 &  600 &   fr &   88.1 &                                                                                                                                                                            Hershey et Chase ont utilisé des phages, ou des virus, pour implanter leur propre ADN dans un bactérie. \\
     7 &  575 &   fr &  100.0 &                                                                                                                                                                                              Mais il y a beaucoup de choses sur les oiseaux qui ressemblent encore à un dinosaure. \\
     \midrule
     8 &  600 &   fr &   85.7 &                                                                                                                                                                            Hershey et Chase ont utilisé des phages, ou des virus, pour implanter leur propre ADN dans un bacillus. \\
     8 &  575 &   fr &  100.0 &                                                                                                                                                                                              Mais il y a beaucoup de choses sur les oiseaux qui ressemblent encore à un dinosaure. \\
     \midrule
     9 &  600 &   fr &   78.8 &                                                                                                                                                                           Hershey et Chase ont utilisé des phages, ou des virus, pour implanter leur propre gène dans un bactérie. \\
     9 &  575 &   fr &  100.0 &                                                                                                                                                                                              Mais il y a beaucoup de choses sur les oiseaux qui ressemblent encore à un dinosaure. \\
    \midrule
    10 &  600 &   fr &   30.1 &                                                                                                                                                                             En gros, vous mettre en place des phages, ou viraux, pour implanter leur propre gène dans un bactérie. \\
    10 &  575 &   fr &  100.0 &                                                                                                                                                                                              Mais il y a beaucoup de choses sur les oiseaux qui ressemblent encore à un dinosaure. \\
    \midrule
    11 &  600 &   fr &    1.3 &  Fondamentalement, vous afficherez des annonces pour proposer votre aide, arpenterez les quais, aborderez les personnes nettoyant leurs yachts, ... \\
    11 &  575 &   fr &    1.4 &                                                                    Fondamentalement, vous afficherez des annonces pour proposer votre aide, arpenterez les quais, aborderez les personnes nettoyant leurs yachts, . \\
    \midrule
    12 &  600 &   fr &   12.4 &                                                                                                                                                                                                            Les phages sont utilisés pour implanter leur propre gène dans un virus. \\
    12 &  575 &   fr &   36.8 &                                                                                                                                                                                                                   Mais il y a beaucoup de choses qui semblent être des dinosaures. \\
    \midrule
    13 &  600 &   fr &    4.3 &                                                                                                                                                                                                       Les phages sont des virus qui sont implantés dans la cellule d'un organisme. \\
    13 &  575 &   fr &    5.5 &                                                                                                                                                                                                               Les oiseaux sont des animaux, mais ils sont aussi des êtres humains. \\
    \midrule
    14 &  600 &   fr &   74.4 &                                                                                                                                                                                Hershey et Chase ont utilisé des phages, ou virus, pour implanter leur propre ADN dans un bactérie. \\
    14 &  575 &   fr &   73.3 &                                                                                                                                                                                                 Mais il y a beaucoup de choses sur les oiseaux qui ressemblent à un dinosauresque. \\
    \midrule
    15 &  600 &   fr &   62.6 &                                                                                                                                                                                Hershey et Chase ont utilisé des phages, ou virus, pour implanter leur propre gène dans un bacille. \\
    15 &  575 &   fr &   26.7 &                                                                                                                                                               Mais il y a beaucoup de choses à propos de la façon dont les oiseaux se ressemblent, même si c'est un peu plus tard. \\
    \midrule
    16 &  600 &   fr &   85.7 &                                                                                                                                                                            Hershey et Chase ont utilisé des phages, ou des virus, pour implanter leur propre ADN dans un bacillus. \\
    16 &  575 &   fr &   76.7 &                                                                                                                                                                                         Mais il y a beaucoup de choses sur les oiseaux qui ressemblent encore comme un dinosaures. \\
    \midrule
    17 &  600 &   fr &   85.7 &                                                                                                                                                                             Hershey et Chase ont utilisé des phages, ou des virus, pour implanter leur propre ADN dans un bacille. \\
    17 &  575 &   fr &   88.4 &                                                                                                                                                                                          Mais il y a beaucoup de choses sur les oiseaux qui ressemblent encore à un dinosauresque. \\
    \midrule
    18 &  600 &   fr &   85.7 &                                                                                                                                                                             Hershey et Chase ont utilisé des phages, ou des virus, pour implanter leur propre ADN dans un bacille. \\
    18 &  575 &   fr &   88.4 &                                                                                                                                                                                          Mais il y a beaucoup de choses sur les oiseaux qui ressemblent encore à un dinosauresque. \\
    \midrule
    19 &  600 &   fr &   76.5 &                                                                                                                                                                            Hershey et Chase ont utilisé des phages, ou des virus, pour implanter leur propre gène dans un bacille. \\
    19 &  575 &   fr &   88.4 &                                                                                                                                                                                               Mais il y a beaucoup de choses sur les oiseaux qui ressemblent encore à un dinosaur. \\
    \midrule
    20 &  600 &   fr &   76.5 &                                                                                                                                                                           Hershey et Chase ont utilisé des phages, ou des virus, pour implanter leur propre gène dans un bacillus. \\
    20 &  575 &   fr &   88.4 &                                                                                                                                                                                               Mais il y a beaucoup de choses sur les oiseaux qui ressemblent encore à un dinosaur. \\
    \midrule
    21 &  600 &   fr &   88.1 &                                                                                                                                                                            Hershey et Chase ont utilisé des phages, ou des virus, pour implanter leur propre ADN dans un bactérie. \\
    
    21 &  575 &   fr &   88.4 &                                                                                                                                                                                               Mais il y a beaucoup de choses sur les oiseaux qui ressemblent encore à un dinosaur. \\
    \midrule
    22 &  600 &   fr &   85.7 &                                                                                                                                                                             Hershey et Chase ont utilisé des phages, ou des virus, pour implanter leur propre ADN dans un bacille. \\
    22 &  575 &   fr &   88.4 &                                                                                                                                                                                              Mais il y a beaucoup de choses sur les oiseaux qui ressemblent encore à un dinosauro. \\
    \midrule
    23 &  600 &   fr &   88.1 &                                                                                                                                                                            Hershey et Chase ont utilisé des phages, ou des virus, pour implanter leur propre ADN dans un bactérie. \\
    23 &  575 &   fr &   88.4 &                                                                                                                                                                                               Mais il y a beaucoup de choses sur les oiseaux qui ressemblent encore à un dinosaur. \\
    \midrule
    24 &  600 &   fr &   85.7 &                                                                                                                                                                             Hershey et Chase ont utilisé des phages, ou des virus, pour implanter leur propre ADN dans un bacille. \\
    24 &  575 &   fr &   88.4 &                                                                                                                                                                                               Mais il y a beaucoup de choses sur les oiseaux qui ressemblent encore à un dinosaur. \\
    \midrule
    25 &  600 &   fr &   76.5 &                                                                                                                                                                           Hershey et Chase ont utilisé des phages, ou des virus, pour implanter leur propre gène dans un bacillus. \\
    25 &  575 &   fr &   88.4 &                                                                                                                                                                                               Mais il y a beaucoup de choses sur les oiseaux qui ressemblent encore à un dinosaur. \\
    \midrule
    26 &  600 &   fr &   76.5 &                                                                                                                                                                            Hershey et Chase ont utilisé des phages, ou des virus, pour implanter leur propre gène dans un bacille. \\
    26 &  575 &   fr &   88.4 &                                                                                                                                                                                               Mais il y a beaucoup de choses sur les oiseaux qui ressemblent encore à un dinosaur. \\
    \midrule
    27 &  600 &   fr &   55.9 &                                                                                                                                                                         Hershey et Château utilisaient des phages, ou des virus, pour implanter leur propre gène dans un bactérie. \\
    27 &  575 &   fr &   88.4 &                                                                                                                                                                                               Mais il y a beaucoup de choses sur les oiseaux qui ressemblent encore à un dinosaur. \\
    \midrule
    28 &  600 &   fr &   76.5 &                                                                                                                                                                           Hershey et Chase ont utilisé des phages, ou des virus, pour implanter leur propre gène dans un bacillus. \\
    28 &  575 &   fr &   88.4 &                                                                                                                                                                                               Mais il y a beaucoup de choses sur les oiseaux qui ressemblent encore à un dinosaur. \\
    \midrule
    29 &  600 &   fr &   78.8 &                                                                                                                                                                           Hershey et Chase ont utilisé des phages, ou des virus, pour implanter leur propre gène dans un bactérie. \\
    29 &  575 &   fr &  100.0 &                                                                                                                                                                                              Mais il y a beaucoup de choses sur les oiseaux qui ressemblent encore à un dinosaure. \\
    \midrule
    30 &  600 &   fr &   78.8 &                                                                                                                                                                           Hershey et Chase ont utilisé des phages, ou des virus, pour implanter leur propre gène dans un bactérie. \\
    30 &  575 &   fr &   88.4 &                                                                                                                                                                                               Mais il y a beaucoup de choses sur les oiseaux qui ressemblent encore à un dinosaur. \\
    \midrule
    31 &  600 &   fr &   78.8 &                                                                                                                                                                           Hershey et Chase ont utilisé des phages, ou des virus, pour implanter leur propre gène dans un bactérie. \\
    31 &  575 &   fr &  100.0 &                                                                                                                                                                                              Mais il y a beaucoup de choses sur les oiseaux qui ressemblent encore à un dinosaure. \\
    \midrule
    32 &  600 &   fr &   78.8 &                                                                                                                                                                           Hershey et Chase ont utilisé des phages, ou des virus, pour implanter leur propre gène dans un bactérie. \\
    32 &  575 &   fr &   57.2 &                                                                                                                                                                                   Mais il y a beaucoup de choses sur les oiseaux que pourraient encore ressembler à un dinosaures. \\
\bottomrule
\end{tabular}
\caption{5-prompts without instructions, masking layer by layer of \textsc{GPTNeo}2.7B}
\label{tab:gen_text_nprompts5_layerwise_noinstr}
\end{table*}
\clearpage




\end{document}